\documentclass[journal]{IEEEtran}
\usepackage{algorithm}
\usepackage{algorithmic}
\usepackage{wrapfig}
\usepackage{helvet}  
\usepackage{courier}  
\usepackage{graphicx}  
\usepackage{epstopdf}
\usepackage{subfigure}
\usepackage{float}
\usepackage{epsfig}
\usepackage{amsmath,amsthm,amssymb,amsfonts}
\usepackage{enumerate}
\usepackage{array}
\usepackage{multirow}
\usepackage{url}
\usepackage[square, comma, sort&compress, numbers]{natbib}

\ifCLASSINFOpdf
\else
\fi
\usepackage{xspace}
\makeatletter
\DeclareRobustCommand\onedot{\futurelet\@let@token\@onedot}
\def\@onedot{\ifx\@let@token.\else.\null\fi\xspace}
\def\eg{\emph{e.g}\onedot} 
\def\ie{\emph{i.e}\onedot}

\makeatother

\usepackage{color}
\definecolor{mygray}{gray}{0.6}
\definecolor{dg}{rgb}{0,0.694,0.298}
\definecolor{purple}{rgb}{0.4,0.176,0.569}
\definecolor{pink}{cmyk}{0, 0.7808, 0.4429, 0.1412}

\begin{document}
%
\title{Independent Reinforcement Learning for Weakly Cooperative Multiagent Traffic Control Problem}
%
%
%


\author{Chengwei~Zhang,
        Shan~Jin,
        Wanli~Xue,
        Xiaofei~Xie,
        Shengyong~Chen,~\IEEEmembership{Fellow,~IET,}
        and~Rong~Chen
\thanks{W. Xue is the corresponding author}
\thanks{C. Zhang, S. Jin and R. Chen are with the School of Information Science and Technology, Dalian Maritime University, Dalian, China. e-mail: ({chenvy,rc}@dlmu.edu.cn).}
\thanks{W. Xue and S. Chen are with Key Laboratory of Computer Vision and Systems (Ministry of Education), School of Computer Science and Engineering, Tianjin University of Technology, Tianjin, China. e-mail: (xuewanli@email.tjut.edu.cn, sy@ieee.org).}
\thanks{X. Xie is with Nanyang Technological University, Singapore. email: xiaofei.xfxie@gmail.com.}
\thanks{This paper is currently submitted in IEEE TRANSACTIONS ON VEHICULAR TECHNOLOGY for reviewing}
}%


\maketitle

\begin{abstract}
The adaptive traffic signal control (ATSC) problem can be modeled as a multiagent cooperative game among urban intersections, where intersections cooperate to optimize their common goal, \ie, the city's traffic conditions. The large scale of intersections in a real traffic scenario yield marked challenges for an algorithm to find an optimal joint control strategy by controlling multiple intersections at the same time. Recently, reinforcement learning (RL) has achieved marked successes in managing sequential decision making problems, which motivates us to apply RL in the traffic control problem. In particular, independent reinforcement learning (IRL) is typically used to solve a multiagent task, where IRL agents learn cooperation strategies independently according to specific rules, such as the `optimistic' or `lenient' principle, and treat other agents as part of the environment.
Considering the large scale of intersections in an urban traffic environment, we use IRL to solve a complex traffic cooperative control problem in this study. One of the largest challenges of this problem is that the observation information of intersection is typically partially observable, which limits the learning performance of IRL algorithms. To mitigate this challenge, we model the traffic control problem as a partially observable weak cooperative traffic model (PO-WCTM) to optimize the overall traffic situation of a group of intersections. Different from a traditional IRL task that averages the returns of all agents in fully cooperative games, the learning goal of each intersection in PO-WCTM is to reduce the cooperative difficulty of learning, which is also consistent with the traffic environment hypothesis.
To determine the optimal cooperative strategy of PO-WCTM, we also propose an IRL algorithm called Cooperative Important Lenient Double DQN (CIL-DDQN), which extends Double DQN (DDQN) algorithm using two mechanisms: the forgetful experience mechanism and the lenient weight training mechanism. The former mechanism decreases the importance of experiences stored in the experience reply buffer, which deals with the problem of experience failure caused by the strategy change of other agents. The latter mechanism increases the weight experiences with high estimation and `leniently' trains the DDQN neural network, which improves the probability of the selection of cooperative joint strategies. Experimental results show that, in two real traffic scenarios and one simulated traffic scenario, CIL-DDQN outperforms other methods in almost all performance indicators of the traffic control problem.
\end{abstract}


%
\IEEEpeerreviewmaketitle

\section{Introduction}
%
%
%
%
\IEEEPARstart{T}{raffic} congestion plagues many cities around the world, producing heavy pressure on existing urban transport infrastructure. Using traffic signals to control vehicle waiting time is one of the most effective ways to relieve congestion. Traditional traffic methods~\cite{liang2019a,gartner1990multiband--a,cools2006self-organizing,luk1984two,hunt1982the}, which select the phase from the predefined signal combination based on expert knowledge, have certain limitations, and cannot adjust traffic based on real-time traffic conditions. Due to the nature of the interaction between the agent and the environment in reinforcement learning (RL), RL can be naturally applied to the traffic signal control environment. Many RL studies~\cite{nishi2018traffic,wei2019colight,zhu2015a} have reported the advantages of RL in traffic control scenarios.

In a real urban traffic environment, there may be thousands of intersections that must cooperate to optimize transportation. An action taken by one intersection affects its local traffic pressure and affects the road conditions at other intersections. The state transition function and reward function of the traffic environment depend on the phase selection of all intersections, making the traffic environment dynamic and unstable. Therefore, how to apply RL in such a complex environment has become a challenge. Specifically, the urban traffic control problem, which aims to minimize the average travel time of vehicles through multiple intersections, can be defined as a multiagent cooperative problem. The direct application of single-agent RL algorithms to the multiagent environment may create new problems, such as the non-stationary problem~\cite{bloembergen2015evolutionary,matignon2012review}, which often leads to non-cooperation or even fail to convergence for an RL algorithm in learning a joint optimal policy of a multiagent reinforcement learning (MARL) environment.

To solve the non-stationary problem, a natural way is to use a centralized training framework that can learn the estimation of an joint action in the global state. Thus, most existing MARL algorithms use centralized training with a distributed execution mechanism to learn the optimal joint strategy~\cite{sunehag2017value-decomposition,lowe2017multi-agent,son2019qtran}. By considering global state information, these methods effectively mitigate the problems of a non-stationary environment and solve the multiagent cooperative problem somewhat accurately. However, in real urban traffic environments, the scale of the road network is often massive. Centralized learning is difficult to scale because the joint action space grows exponentially as the number of agents increases linearly. In contrast, independent reinforcement learning (IRL) is more flexible and suitable for traffic control because it is not constrained by the number of agents, where each intersection can be controlled by a local RL agent. Besides, it is often unrealistic to collect the global state information and feed them to the agent in a traffic environment. Usually, the state information of the agent in the traffic environment is the partial observability and/or the communication constrained. 
Independent RL learners treat other agents as part of the environment and train their networks independently using a predetermined principle (\eg, the `optimistic' principle~\cite{wei2016lenient}). They update the evaluation of action only if the new evaluation is greater than the previous one (or if they prefer the new evaluation) by identifying and discarding experience (HDQN~\cite{omidshafiei2017deep} and LDQN~\cite{palmer2018lenient}) or trajectories (IGASIL~\cite{HaoWHY19} and NUI-DDQN~\cite{palmer2019negative}).

However, the importance of experience (or trajectories) is difficult to identify, especially in games where the optimal policy of the game is much more difficult to explore than sub-optimal policies. The aforementioned methods make strong assumptions about the environment, \ie, games are required to be fully cooperative and small in scale. Thus, these methods cannot be directly applied to urban traffic scenarios for two reasons.
First, traffic conditions in different areas of a city are often different, which makes it unnecessary to model the traffic control problem as a fully cooperative problem. Our experimental results show that it actually could be counterproductive to define traffic as a fully cooperative problem. Second, intersections that are adjacent to a given intersection have a greater impact on congestion at that intersection than at a distant intersection. The state and action information of those adjacent intersections is also typically observable in the intersection. That information is thus useful when optimizing traffic conditions and should not only be treated as environmental components.


In this study, we first modeled the urban traffic control problem as a modified cooperative Markov game, which we refer to as the \textit{partially observable weakly cooperative traffic model} (PO-WCTM). Unlike the average return of all agents used in fully cooperative Markov games in existing IRL studies~\cite{omidshafiei2017deep,palmer2018lenient,palmer2019negative}, the learning goal of each agent (intersection) in PO-WCTM is defined as the average return within its local scope, \ie, the average return of intersections of the agent and its neighbors, so that the intersection can pay more attention to the traffic conditions in its vicinity. Besides, the observable state information of each intersection is also defined as information within its local scope, which is in line with the assumption of real traffic environment.
This design is also convenient for the MARL algorithm during development and training. Conversely, collecting and processing the global state will increase the time delay and cost of the system, which may affect training performance.
Based on PO-WCTM, we thus propose an IRL method called the \textit{Cooperative Important Lenient Double DQN} (CIL-DDQN) method. Specifically, CIL-DDQN adds two cooperative mechanisms to the Double DQN (DDQN)~\cite{van2016deep} method: the optimistic weigh training mechanism and the forgetful experience mechanism. The first mechanism improves the leniency mechanism used in LDQN~\cite{palmer2018lenient} to weight the empirical importance according to the TD error of sampled data during the training process. By regulating `leniency', which is a parameter that decreases as the training time increases to adjust the degree of `optimism', during the negative update, the new loss function makes the agent prefer optimistic experience training, increasing the possibility of reaching the cooperative joint strategy. The forgetful experience mechanism is proposed to mitigate the limitation of the partially observable experience that is stored in the experience reply memory (ERM). The data stored in the early stage cannot reflect the current situation of training if strategies of other agents are changed. The forgetful experience mechanism decreases the importance of experiences stored in ERM as their storage time increases. Based on the two experience weight allocation mechanisms, the training performance of CIL-DDQN is markedly improved.

The performance of the proposed algorithm is evaluated with two real traffic datasets and one synthetic dataset. Experimental results show that CIL-DDQN outperforms other traffic control methods (\eg, MA2C~\cite{chu2020multi-agent} and Colight~\cite{wei2019colight}) in almost all criteria, including queue length, intersection delay, vehicle speed, trip completion flow, and trip delay, which are commonly used to evaluate traffic conditions.

The reminder of this paper is organized as follows. A literature review is presented in Section \ref{sec:related work}. Section \ref{sec:background} introduces relevant background on RL and notations used in traffic control scenarios. The proposed PO-WCTM model and problem statement are introduced in Section \ref{sec:model}, followed by the proposed independent reinforcement learning method CIL-DDQN in Section \ref{sec:mehthod}. The model and algorithm are evaluated in Section \ref{sec:experiment}, and the paper is concluded in Section \ref{sec:con}.

\section{Related work}\label{sec:related work}
Early research on traditional traffic signal control methods~\cite{gartner1990multiband--a,cools2006self-organizing,luk1984two,hunt1982the} mostly relied on the expert strategy, which simplifies the traffic conditions. The performance of previous traffic detection equipment limits the collection of traffic data, and affects the effect of traditional traffic control methods. With the development of the machine learning theory, deep reinforcement learning (DRL) has become a new trend in adaptive signal traffic control. To reduce potential congestion, adaptive methods adjust signal timing according to real-time traffic dynamically. RL is defined to find the optimum strategy of an unknown environment with sequential decision making problems by trial and error. According to the intersections scale, traffic control studies based on RL can be divided into two categories: single-intersection control and multiple-intersections control.

\textbf{Single-intersection control}.  Wiering et al.~\cite{Wiering2004Intelligent} first used tabular Q-learning to the single-intersection control problem. With the development of detection facilities, the number of detectable features of traffic state information (\eg, vehicle position, speed, length of road queue) increased, which exceeded the capability of tabular Q-learning. With the development of deep learning and reinforcement learning, new methods that these two processes into DRL have been used to estimate Q. Li et al.~\cite{li2016traffic} considered traffic-like information as the state input and estimated the Q function using a deep neural network (DNN) by DQN~\cite{Mnih2015Human}. 
Because traffic states are sequential, and there is a causal relationship between them, Choe et al.~\cite{choe2018deep} analyzed the sequence's state information through a recurrent LSTM network with deep recursive Q learning. In addition to designing different networks, some researchers have applied DQN enhancement mechanisms to traffic.
Liang et al.~\cite{liang2019a} integrated optimization factors including duling DQN~\cite{wang2016dueling}, DDQN~\cite{van2016deep}, priority experience replay, and a convolutional neural network to map the state to reward.


\textbf{Multiple-intersections control}.  There are also studies that investigated controlling multiple-intersections within a given area of a city. 
Zhu et al.~\cite{zhu2015a} proposed a joint tree in the probability graph model to simplify the calculation of joint action probabilistic reasoning. Wei et al.~\cite{wei2019colight} used a graph attention mechanism to express the effect of other intersections explicitly on the target intersection, to communicate between intersections. Those studies used a centralized training network to estimate the return of a joint policy of all intersections. Centralized methods can help to avoid the non-stationary problem and achieve good performance in multiagent learning environments; however, they are strongly affected by the number of learning agents. Independent learners control intersections can be expanded and trained. Nishi et al.~\cite{nishi2018traffic} combined a graph convectional network with a traffic road network. Intersections in the road network were regarded as nodes of the adjacency matrix, and an oriented graph of traffic flow was constructed to describe traffic conditions in detail and accurately. Chu et al.~\cite{chu2020multi-agent} proposed a scalable algorithm called MA2C, which stabilizes the independent A2C algorithm by adding neighbor information and neighbor strategy in the state, which is equivalent to local communication between the target intersection and the neighbor intersection. However, this algorithm only increased the coordination between intersections from the perspective of traffic models.

Most existing methods extend the single-agent RL methods directly to a traffic environment. Because the agent strategy in a multiagent environment is non-stationary where agents are faced with changing target decision issues, these algorithms cannot learn good cooperation strategies in a complex road network environment.

\section{Notation and background}\label{sec:background}
This section briefly describes some basic notations of traffic signal control problems and MARL.

\subsection{Notation of Traffic Control Problem }
\begin{figure}
\centering
\subfigure[Intersection]{
\label{fig:intersection}
\includegraphics[width=4.5cm]{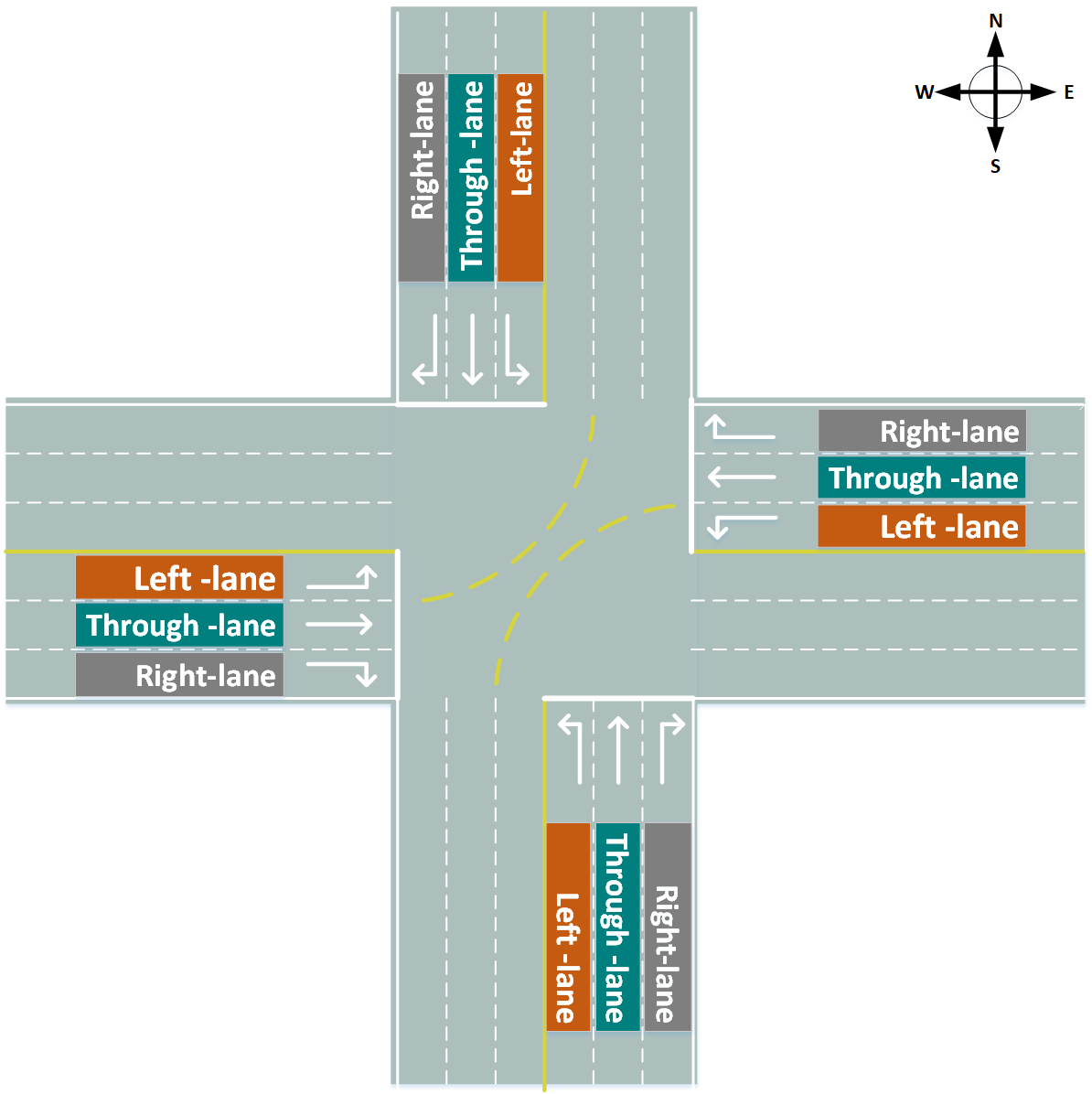}
}
\hfill
\subfigure[Eight phases]{
\label{fig:phase}
\includegraphics[width=2.5cm]{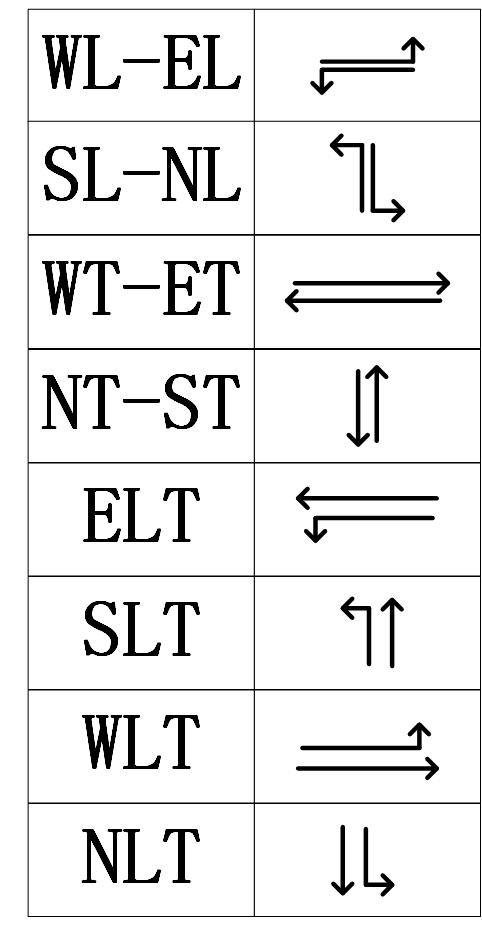}
}
\label{fig:intersection and phase}
\caption{ Definitions of incoming lane and signal phases}
\end{figure}

\textbf{Incoming lane}: The lane where vehicles can enter the intersection is the incoming lane of the intersection. There are 12 incoming lanes at the intersection in Fig.\ref{fig:intersection}. $L_{i}$ represents the set of incoming lanes of intersection $i$. There are four incoming directions at the intersection: West (W), East (E), North (N), and South (S).

\textbf{Signal phase}: The combination of traffic signals in each incoming lane describes a phase. A green signal allows cars to pass. A red signal prohibits cars from passing. Traffic signals can indicate three traffic movements: Left (L), Through (T), and Right (R). Usually, a right-turn vehicle moves through the intersection regardless of the signal. Fig.\ref{fig:phase} shows eight non-conflicting signal phases.


\textbf{Neighbors of intersection}
An urban traffic system consists of multiple connected intersections, where the relationship between intersections can be defined as a graph $G(\mathcal{N}, \mathcal{E})$ according to their geographical location, in which $i \in \mathcal{N}$ is each intersection and $ij \in \mathcal{E}$ represents the road between two intersections $i,j \in \mathcal{N}$. We denote $\mathcal{N}_i = \{ j \in \mathcal{N} | ij \in {\mathcal{E}}\}$ as the neighbors of intersection $i$.

\subsection{RL and MARL}
 \textbf{Partially Observable Markov Game}: We consider the standard MARL setting, in which the interaction of agents and environment is modeled as a Markov game, where multiple agents make choices sequentially. 
Formally, a Markov game is defined by a tuple $G=\left\langle {\mathcal{N},\mathcal{S},\mathcal{O},\mathcal{A},P,R,\gamma}\right\rangle$. $\mathcal{N}$ is the set of $N$ agents, \ie, $|\mathcal{N}|=N$.
$\mathcal{S}$ is the set of states and $\mathcal{O}=\left\langle {\mathcal{O}_1,...,\mathcal{O}_N}\right\rangle $ is the observation set, where $\mathcal{O}_i$ is the observation set of agent $i$. $\mathcal{A}=\left\langle {\mathcal{A}_1,...,\mathcal{A}_N} \right\rangle$ is the set of the joint actions, where $\mathcal{A}_i$ is the set of the action for agent $i$. $P:S \times \mathcal{A} \times \mathcal{S} \to [0,1]$ is the transition function returning the probability of transitioning from a state $s$ to $s'$ given a joint action $\left\langle {{a_1},...,{a_N}} \right\rangle$. $R=\left\langle {{r_1},...,{r_N}} \right\rangle$ is the reward function, where ${r_i}:S \times \mathcal{A} \to \mathbb{R}$ specifies the reward for agent $i$ given the state and the joint action. $\gamma$ is the discount factor. A Markov game is a team game if every player gets the same reward. 
Thus, team games are fully cooperative settings, where players have a shared objective.

The policy $\pi_i$ of agent $i$ represents a mapping from the observation space to a probability distribution over actions: ${\pi_i}:{O_i} \to \Delta ({A_i})$, while $\pi=\left\langle \pi_i,\pi_{- i} \right\rangle$ refers to a joint policy of all agents, and $\pi _{ - i}$ is the joint policies excluding agent $i$. Given a joint policy $\pi$ the return (or expected sum of future rewards) for each agent $i$ starting from a state $s$ can be defined by 
the state-value function, which is also known as $Q$ value function, where $r_{i,t}$ refers to the reward received by agent $i$ at time $t$: $Q_{i,\pi} (s,a) =  \mathbb{E}_\pi [ \sum_{k = 0}^\infty {{\gamma ^k}r_{i,t+k+1}|{s_t} = s, {a_t} = u} ]$.

For a Markov game, a joint policy $\pi^*$ is a Nash equilibrium (NE) if and only if no agent can improve it’s gain through unilaterally deviating from $\pi^*$. 
From a group perspective, NE is often sub-optimal. In contrast, Pareto-optimality defines a joint policy $\hat \pi$ from which no agent can deviate without making at least one other agent worse. 
A NE joint policy $\hat \pi^*$ is Pareto-optimal if it is not Pareto-dominated by any other NE. In the cooperative multiagent learning literature, especially for a team game, convergence to Pareto optimal NE is the most commonly accepted goal to pursue, 
considering that multiple players cooperate to maximize their goals~\cite{Matignon2012,palmer2018lenient}.

\textbf{Deep Q-Learning and DDQN}: The combination of deep neural networks and RL completes the direct mapping of the state to the Q function. Deep Q-learning represents the action-value function with a deep neural network parameterized by $\theta$. Deep Q-Networks (DQNs)~\cite{Mnih2015Human} use a \emph{replay memory} to store the experience $\left\langle {s,a,r,s'} \right\rangle$, where $s'$ is observed after taking the action $a$ in state $s$ and receiving reward $r$. $\theta$ is learnt by sampling batches of $b$ experiences from the replay memory and minimising the squared TD error:
\begin{equation}
\label{eq:func_1}
\begin{split}
L_{DQN}(\theta ) = \sum\limits_{i = 1}^b {[{{({y_i} - Q(s,a;\theta ))}^2}]}
\end{split}
\end{equation}
where learning target $y = r + \gamma {\max _{a'}}Q(s',a';{\theta ^ - })$. $\theta ^ - $ are the parameters of a target network that are periodically copied from $\theta$ and kept constant for a number of iterations.

DQN is affected by an overestimation bias due to the maximization step in Eq.\ref{eq:func_1}, which can harm learning. DDQN~\cite{van2016deep} addresses this overestimation by decoupling the selection of the action from its evaluation during the maximization performed for the bootstrap target using a refined loss $L_{DDQN}(\theta)$, whose learning target is replaced by
\begin{equation}
\label{eq:func_2}
\begin{split}
y= r + \gamma Q( {s',\mathop {{\mathop{\rm argmax}\nolimits} }\limits_a Q\left( {s',a;{{\bf{\theta }}}} \right);{\bf{\theta }}^ - } )
\end{split}
\end{equation}


\textbf{Lenient Deep Q-Network(LDQN)}~\cite{palmer2018lenient}: The agent’s early exploration of the action’s rewards is typically worse. The lenient agent forgives the punishment caused by teammates at the beginning of training, which increases the possibility of cooperation between independent learners~\cite{claus1998the}, \ie, agents that treat other agents as part of the environment and make decisions based on their local observations, actions and rewards only. Each state-action pair explored by the agent has a corresponding temperature value, and the initial temperature is the highest. When the same state-action pair is accessed again, the corresponding temperature of the state-action pair decreases. The leniency function is as follows:
\begin{equation}
\label{eq:func_3}
\begin{split}
l({s_t},{a_t}) = 1 - {e^{ - K \times T({\phi (s)},a)}}
\end{split}
\end{equation}
where $K$ is a leniency moderation factor, and $\phi (s)$ is a hash-key to encode the state. The leniency is directly proportional to the temperature drop of the state-action pair. Therefore, as the number of training steps and frequent access to the same state-action pair increase, the lenient agent will strictly deal with low return data. The experience replay memory (ERM) stores the amount of leniency and transitions experienced by the agent. According to the leniency value of the sampling data, the data is partially discarded when Q is updated. Given TD-error $\delta = {Y_{\rm{t}}} - {Q_t}({s_t},{a_t};\theta )$, the Q update function is as follows:
\begin{equation}
\label{eq:func_4}
\begin{split}
Q({s_t},{a_t}) = \left\{ {\begin{array}{*{20}{l}}
{Q({s_t},{a_t}) + \alpha \delta }&{\delta > 0 \vee  x > l({{\rm{s}}_t}{\rm{,}}{{\rm{a}}_t}{\rm{) }}}\\
{Q({s_t},{a_t}){\rm{ }}}&{ \delta  \le 0 \vee x \le {\rm{l(}}{{\rm{s}}_t}{\rm{,}}{{\rm{a}}_t}{\rm{)}}}
\end{array}} \right.
\end{split}
\end{equation}
where $x\sim U\left( {0,1} \right)$ is uniformly distributed.

\section{MARL MODEL OF TRAFFIC SIGNAL CONTROL}\label{sec:model}
This study investigates the control of a group of intersections in a city to optimize the overall traffic state of the city. Considering that traffic conditions of multiple intersections of a city typically affect each other, directly applying single RL methods to a multi-intersection environment
may yield poor performance, which will be shown described later. Also, the influence of different intersections could be different. For example, congestion at an intersection is more likely to be affected by its adjacent intersections rather than other intersections that are farther away. Traffic conditions in different areas of a city are frequently often different, which makes it unnecessary to model the traffic signal control problem as a fully cooperative problem. Defining a traffic problem as a fully cooperative problem may be counterproductive.

Thus, we model the urban traffic control problem as a modified cooperative Markov game, which we call the partially observable weakly cooperative traffic model (PO-WCTM). Unlike the average return of all agents used in fully cooperative Markov games in existing IRL studies, the learning goal of PO-WCTM is defined as the average return of intersections of the target intersection and its neighbors, so that intersections can pay more attention to the traffic conditions within its vicinity. In addition, the observable state information of each intersection is also defined as information within its local scope, which is in line with the assumption of a real traffic environment. Considering the computation cost of global state collection and process that may affect training performance, the proposed design is more convenient with the MARL algorithm in terms of realization and training. We will introduce the definition of the three key elements of PO-WCTM: observations, actions, and rewards.

\subsection{Observations}
An IRL agent cannot observe the entire environment but rather part information of the environment $o$~\cite{zheng2019diagnosing}. We define the local observation of an intersection as its current phase and the total number of approaching vehicles along each incoming lane:

\begin{equation}
\label{eq:func_5}
\begin{split}
{{\rm{o}}_{t,i}} = ({phase_{\rm{t,i}}},wave_{\rm{t}}{\left[ l \right]_{l \in {L_i}}})
\end{split}
\end{equation}
where $phase_{\rm{i}}$ is the current phase of intersection $i$, $l \in {L_i}$ is the incoming lane of intersection $i$ and $wave\left[ l \right]$ measures the total number of approaching vehicle alone incoming lane $l$ of intersection $i$.

\subsection{Actions}
For each intersection, we define the local action as a possible phase, \ie, red-green combinations of traffic lights, see Figure \ref{fig:phase}. We thus consider four red-green combinations of traffic lights: go straight in the north-south direction (NT-ST), go straight in the east-west direction (ST-WT), turn left in the north-south direction (SL-NL), and turn left in the east-west direction (WL-EL).

\subsection{Rewards}
In a fully cooperative multiagent problem, the reward received by each agent is the global reward, \ie, all agents share the same reward, and finding the global optimal strategy is the common goal of all agents. However, it is inappropriate for multiple intersections in an urban traffic scenario to pursue the global optimal return due to the Spatio-Temporal characteristic of urban traffic scenario~\cite{chu2019multi}, where the traffic congestion of an intersection is usually only affected by the actions of the intersection and its adjacent intersections, and each intersection performs distributed control based on its local observations and messages from its connected neighbors. Rewards of intersections far away can interfere with learning, which yields poor results and can hinder convergence during learning. In this study, we define the reward of intersection $i$ as the total length of waiting vehicles on each incoming lane ${l \in L_{\rm{i \cup {\mathcal{N}_i}}}}$ of agent $i$ as well is its neighbors $\mathcal{N}_i$:
\begin{equation}
\label{eq:func_6}
\begin{split}
{r_{t,i}} =  - \frac{{\sum\nolimits_{l \in {L_{i \cup {\mathcal{N}_i}}}} {wait_{\rm{t}}[l]} }}{{1 + |{\mathcal{N}_i}|}}
\end{split}
\end{equation}
where $i\cup {\mathcal{N}_i}$ is the set of intersection $i$ and its connected neighbors, $1+|{\mathcal{N}_i}|$ is the size of the set, and $wait[l]$ measures the number of waiting vehicles alone incoming lane $l$ of intersection $i$. Specifically, the reward of agent $i$ is the average value of waiting vehicles on incoming lanes of all intersections in the set $i\cup{\mathcal{N}_i}$. Because the agent considers the fewest waiting vehicles in the lane, the reward is set to a negative value.

\section{Cooperative Important Lenient Double DQN}\label{sec:mehthod}
This section introduces the proposed CIL-DDQN algorithm in this study, where multiple intersections in an urban traffic scenario are considered cooperatively to optimize the overall traffic situation of the city by independent reinforcement learning. Specifically, we modeled each intersection as a CIL-DDQN agent, which treats other agents as part of the environment and makes decisions based on their local observations, actions, and rewards as defined in the previous section. Different from centralized training methods, cooperative independent learners in the MARL literature must overcome a rich taxonomy of learning pathologies to converge upon an optimal joint-policy, such as non-stationary and alter-exploration problems~\cite{palmer2019negative,Matignon2012}. The former problem is caused by dynamic changes in multiagents strategies, which results in the loss of Markov assumption from the individual perspective of an agent. The later problem is caused by the exploration-exploitation trade-off required by RL, in which the probability of the global exploration of all agents approaches 1 as the number of agents increases, resulting in agents converging upon a sub-optimal joint policy because exploration can lead to penalties. To address these problems, independent learners can use the rate of excellent experience (the `optimistic' principle in HDQN~\cite{omidshafiei2017deep}), the impact of bad experiences can be reduced during training  (`lenient' principle in LDQN~\cite{palmer2018lenient}) to improve the probability of learning joint optimal strategy. Here CIL-DDQN adds two experience weight allocation mechanisms in DDQN to achieve this requirement, \ie, the forgetful experience mechanism and the lenient weight training mechanism.


\subsection{Forgetful experience mechanism}
ERM is an indispensable key to the success of a single-agent RL algorithm. Single-agent RL algorithms cannot typically solve independent learners cooperative problems due to sample inefficiency, which is caused by obsolete experiences stored inside ERM that become inefficient because the policies of other agents change. 
Due to environment instability, independent learners must evaluate the `importance' of experiences stored in ERM. With agents updates, the environment also changes and the importance of experience stored at the beginning of training is different from that at the later stage of training; randomly sampled data no longer reflects the situation of the agent during training. Thus, stored data will gradually become obsolete as the number of iterations increases. 
Data collected earlier by the agent may not fully conform to the situation that the agent is currently facing, which may decrease training performance. In this study, experience stored in ERM of CIL-DDQN is weighted according to its storage time to reduce the importance of earlier experience. Specifically, the experience stored in ERM at time $t$ is defined by
$\left\langle {{\rm{o}_t},{a_t},{r_t},{o_{t+1}},{e_t} } \right\rangle $, where $e_t$ indicates the importance of this experience for training. Importance $e$ of each experience stored in ERM is initialized by $1$ and decreased by a decay rate $d_e$ when an episode is terminated.

\subsection{Lenient weight training mechanism}
When a target moves in a multiagent environment, we regulate the optimistic degree of independent learners to achieve coordination among agents by comprehensively considering the key ideas in HDQN~\cite{omidshafiei2017deep} and LDQN~\cite{palmer2018lenient}. CIL-DDQN relaxed the restriction of LDQN on the association of lenient with each state action, using a gradually decreased lenient setting which is similar to decreasing $\epsilon$ in the $\epsilon$-greedy strategy used in exploration. Then, CIL-DDQN uses two learning rates ($\alpha_a$ and $\alpha_b$) to describe the increase and decrease of $Q_i$-values, respectively, which is similar to HDQN. The loss function for the training of agent $i$ is:
\begin{equation}
\label{eq:func_7}
\begin{split}
{\rm{Loss = }}\frac{1}{{|{\mathcal{B}}|}}\sum\limits_{x \in {\mathcal{B}}} {\hat\delta _x^2}
\end{split}
\end{equation}
where $\mathcal{B}\in D_i$ is a mini-batch randomly sampled from ERM of agent $i$, and $\hat\delta _x$ is the refined TD-error of the experience $x$ as follows:
\begin{equation}
\label{eq:func_8}
\begin{split}
{{\hat \delta }_x} = \left\{ {\begin{array}{*{20}{c}}
{{e_x}{\delta _x}}&{{\delta _x} > 0}\\
{(1 - l_t){e_x}{\delta _x}}&{{\delta _x} \le 0}
\end{array}} \right.
\end{split}
\end{equation}
where $l_t$ is the valuable of lenient at time $t$,  $\delta _x = {r_x} + \gamma Q(o_x^{'},\mathop {{\rm{argmax}}}\limits_a Q( {o_x^{'},a;{\theta _i}} );\theta _i^ - ) - Q( {o_x}, {a_x}; {\theta _i})$ is the TD-error of $x$ calculated by the DDQN schedule.

At the beginning of the exploration, penalties are ignored because most selected actions are poor choices. Nevertheless, this may lead to an overestimation of actions, especially in stochastic domains where rewards are noisy. As training times increase, the algorithm must also ensure that it can converge to a stable strategy. Thus, agents are initially lenient (or optimistic), and the degree of leniency decreases as the training times increase. The importance $e_x$ of data $x$ mitigates the divergence of sampling data in the memory according to their storage time. The full algorithm CIL-DDQN is shown in Algorithm \ref{algorithm:1}. For each agent $i$, it maintains two randomly initialized neural networks, a DDQN network $\theta_i$  and a target network and $\theta_i^{-}$. During the learning periods, agents choice their actions using the $\epsilon$-greedy strategy (line 5), where $\epsilon$ is gradually decreased from 1 to 0 during the learning process to balance the exploration and exploitation. After all agents execute their actions, each of them will abstain a reward $r_i$ and a new observation $o'_i$ (line 6). Together with an importance $e_i=1$, the new experience is stored into the REM of the agent (line 7). Then the network $\theta_i$ is trained according to Eq.\ref{eq:func_7} with a randomly sampled mini-batch (line 8). The target network $\theta_i^{-}$ is updated softly, where $\tau$ is the soft update weighting factor for target network updating (line 9). Finally, CIL-DDQN updates all other parameters.

\begin{algorithm}[tb]
\centering
\caption{CIL-DDQN for agent $i$}
\label{algorithm:1}
\begin{algorithmic}[1]
\STATE {Initialize evaluation network and target network with parameters $\theta_i$ and $\theta_i^{-}=\theta_i$;}
\STATE {Initialize ERM $D_i$, parameter $l$, $\tau$ and $\epsilon$.}
\FOR{episode=1,...,$M$}
\FOR{$t\leq T$ and not terminal}
\STATE Observe $o_{i}$ and selects an action $a_i$ by $\theta_i$ with $\epsilon$-greedy strategy.
\STATE Observe its reward $r_{i}$ and next observation $o_i'$ after all agents execute their actions.
\STATE Store transition $(o_{i},a_{i},r_{i},o_{i}',e_i=1)$ in $D_i$.
\STATE Update evaluate network $\theta_i$ with a randomly sampled mini-batch from $D_i$ by Eq.\ref{eq:func_7}.
\STATE Update the target networks:
    $\theta_i^{-}=\tau\theta_i+(1-\tau)\theta_i^{-}$.
\STATE Update $l$ and $\epsilon$ by the decay rate $d_l$ and $d_{\epsilon}$ respectively.
\ENDFOR
\STATE {Update $e_i$ for all experience in $D_i$ by $e_i \leftarrow d_{e}e_i $.}
\ENDFOR
\end{algorithmic}
\end{algorithm}


\subsection{Algorithm Analysis}

\begin{figure}
\centering
\includegraphics[width=50mm]{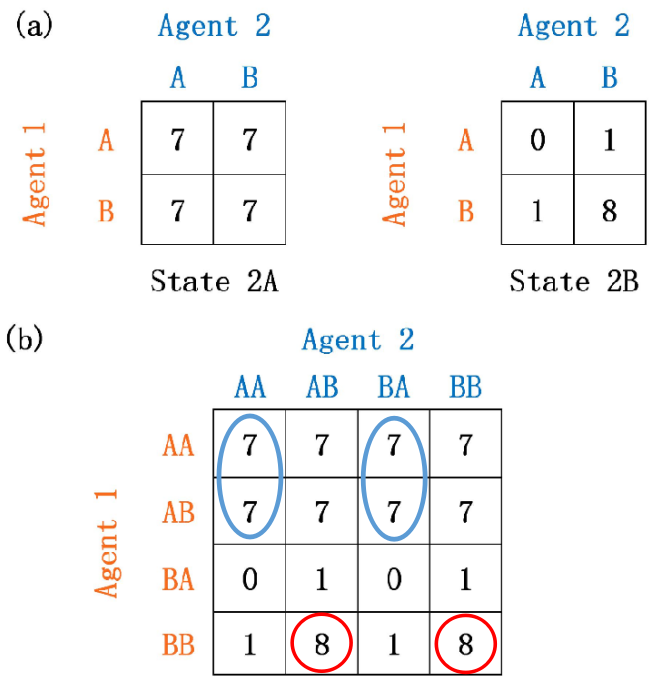}
\caption{(a) Payoff matrices of the two-step game after Agent 1 chose the first action. Action A takes the agents to State 2A and action B takes them to State 2B; (b) The equivalent matrix game of the two-step game. The elements marked by the circle are NEs, among which those marked by the red circle are Pareto-optimal NEs.}
\label{fig:two-step game} 
\end{figure}

\textbf{Performance Analysis}: To explain why CIL-DDQN can learn Pareto-optimal NE, we use a simple two-step cooperative matrix game to illustrate it in detail. The game is a recognized game to analyze cooperative RL algorithms proposed in Qmix~\cite{rashid2018qmix}. At each time step, the two agents make a decision among two actions A and B. At the first step, Agent 1 chooses which of the two matrix games to play in the next time step. For the first time step, the actions of Agent 2 have no effect. In the second step, both agents choose an action and receive a global reward according to the payoff matrices depicted in Figure \ref{fig:two-step game}(a). To make it easy to understand, we transform the game into an equivalent matrix game by combining the actions of each agent in their two steps. Figure \ref{fig:two-step game}(b) shows the payoff matrices of the equivalent matrix game, where each element of the matrix is the payoff of the two agents when they choose the joint action.

\begin{figure}
\centering
\includegraphics[width=65mm]{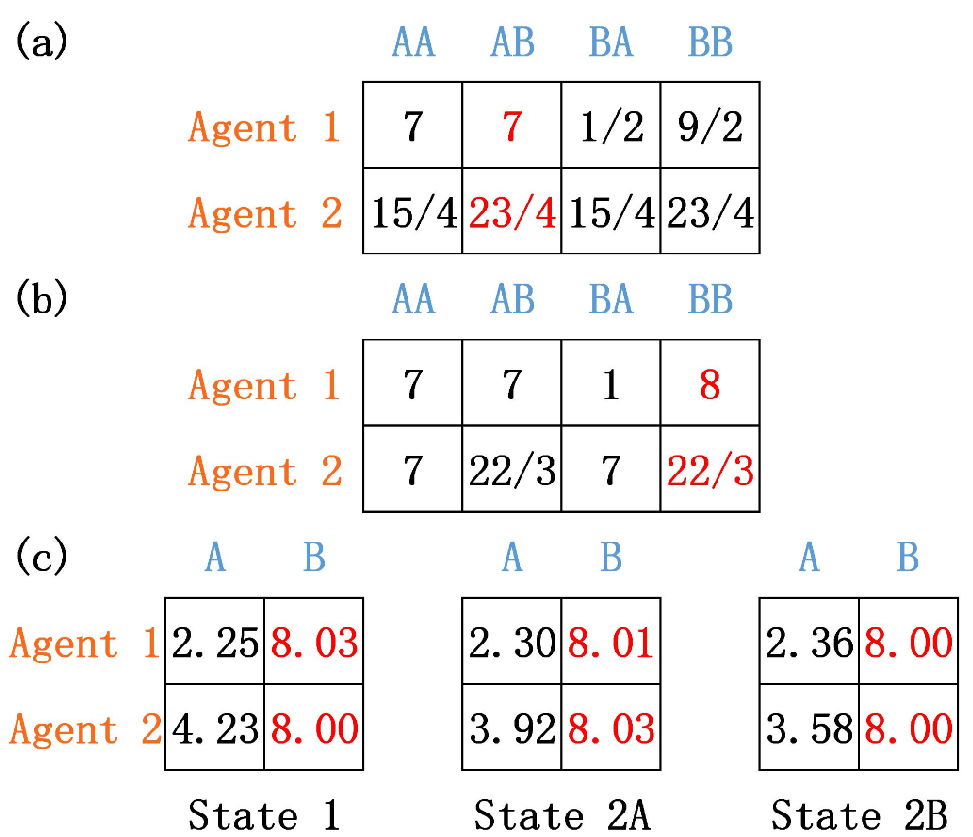}
\caption{(a) Estimated values of the two IDDQN agent in the two-step game; (b) Estimated values of two lenient agents; (c)The learned Q values for the two CIL-DDQN agents. Values marked by red in each table are the greedy values of each agent.}
\label{fig:Q-two-step game} 
\end{figure}

Assuming that the experiences of the two independent learners are gained in the limit of full exploration ($\epsilon = 1$). Full exploration ensures that agents are guaranteed to eventually explore all available game states.To demonstrate, we formulated an independent DDQN (IDDQN) algorithm, by extending the idea of independent Q learning (IQL)~\cite{tan1997multi-agent} on DDQN. For a two-IDDQN agents setting, where all experiences are trained with equal weight, the estimated state-action value of the two IDDQN agents should be the averages in the direction of row and column respectively. Figure \ref{fig:Q-two-step game}(a) shows theoretically estimated state-action values $Q_i(a)$, $a\in \{AA,AB,BA,BB\}$ of agents $i \in \{1,2\}$ in each state. From the figure, we can see that if the two IDDQN agents use greedy policy to interact with the environment, then the joint action of the two agents will be $\left\langle {{\rm{AB}},{\rm{AB}}} \right\rangle $, which is a non-Pareto-optimal NE. The proposed CIL-DDQN treat different experiences differently, where experiences with returns less than average are given lower weights. Figure \ref{fig:Q-two-step game}(b) shows the theoretically estimated state-action values when ignores experience with less than average returns, where the joint greedy action of the two agents are Pareto-optimal NE. We also train CIL-DDQN on this task for 5000 episodes and examine the final learned value functions. The full details of the architecture and hyper-parameters used are shown in section \ref{parameter}. Figure \ref{fig:Q-two-step game}(c) shows the learned values for the two CIL-DDQN agents at each state in the two-step game, demonstrate that the CIL-DDQN is feasible in finding the Pareto-optimal NEs.

\textbf{Space and Time Complexity}:
For the algorithm complexity, there are natural advantages for IRL methods, for the reason that information of other agents are not considered. CIL-DDQN is build based on DDQN, using fully connected neural network, and add some linear operations. Compared with the DDQN, CIL-DDQN only adds a one-dimensional feature for each experience that needs to be stored in ERM, so the CIL-DDQN will not significantly increase the space complexity. In terms of the time complexity, CIL-DDQN trains its network based on the refined TD-error (Eq.\ref{eq:func_8}), where an error judgment is added before calculating the normal DDQN error. Besides, the algorithm has an update operation to refresh the importance value of experiences stored in the ERM. Similarly, it does not significantly increase the computational complexity compared with DDQN.

\section{Experiment} \label{sec:experiment}
We implemented the proposed PO-WCTM model and the CIL-DDQN algorithm using the Cityflow~\cite{zhang2019cityflow} simulator, a commonly used microscopic traffic simulator, in a synthetic traffic grid and two real-world traffic networks in Jinan and Hangzhou city. We first tested CIL-DDQN and other methods ( MA2C~\cite{chu2020multi-agent} and Colight~\cite{wei2019colight}). We also tested IDDQN (independent learners with DDQN) and LDQN~\cite{palmer2018lenient} for ablation of CIL-DDQN, and two traditional traffic methods, (Fixedtime~\cite{gartner1990multiband--a} and SOTL~\cite{cools2006self-organizing}), as in the MA2C paper, for completeness. Experimental results show that CIL-DDQN outperforms the other methods in all metrics. Finally, we tested the proposed PO-WCTM model in three reward setting: local reward (\ie, the reward setting in PO-WCTM), global reward (\ie, the average reward of all agents), and discount reward (\ie, the weighted average of rewards of all agents according to their mutual distances, which is similar with MA2C~\cite{chu2020multi-agent}). Both CIL-DDQN and IDDQN were applied to all three traffic networks. Results show that PO-WCTM is more suitable to model cooperation problems in an urban traffic environment. All source code in this study is available on GitHub \footnote{\url{https://github.com/zcchenvy/CIL-DDQN}}.

\subsection{Traffic scenarios setting}

\begin{figure}
\centering
\subfigure[Dongfeng Street, Jinan]
{\includegraphics[width=34mm]{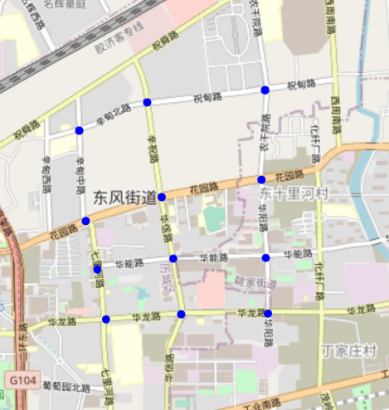}\label{fig:Jinan environment}}
\hfill
\subfigure[Gudang Street,Hangzhou]{\includegraphics[width=50mm]{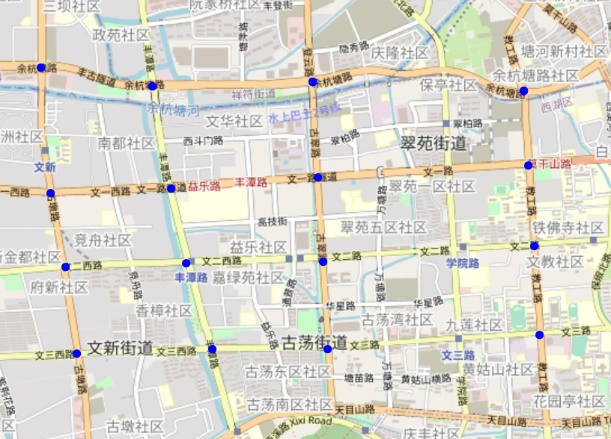}\label{fig:Hangzhou environment}}
\caption{Two real traffic networks of China}
\label{fig:environment} 
\end{figure}

We now investigate three traffic signal control scenarios\footnote{Public datasets are available at \url{https://traffic-signal-control.github.io}}, a $4\times 4$ synthetic traffic grid and two real-world traffic networks from Jinan city (12 intersections) and Hangzhou city (16 intersections) in China, using the traffic simulator Cityflow~\cite{zhang2019cityflow}. The two real-world traffic networks are imported from OpenStreetMap\footnote{\url{https://www.openstreetmap.org}}, as shown in Figure \ref{fig:Jinan environment} and Figure \ref{fig:Hangzhou environment}. All three grids above are homogeneous, and all agents have the same action space, which is a set of four pre-defined signal phases. In all scenarios, each episode simulates peak-hour traffic, and a 10s control interval is used to prevent traffic lights from switching too frequently, based on RL control latency and driver response delay. Thus, one MDP step corresponds to 10s in the simulation, and the horizon is 360 steps.

Vehicle data of the three traffic scenarios contain detailed vehicle information in the road network within an hour, including the starting position, departure time, and driving route of each vehicle. In the synthetic scenario, the number of vehicles generated in each lane is sampled by Gaussian distribution. All vehicles enter the road network from one of an edge node and pass through various intersections before leaving the network. The turning ratio of those generated vehicles during driving is set as 10\% left turn, 60\% straight ahead, and 30\% right turn. For the two real-world scenarios, vehicle data was gathered by cameras at each intersection in the real traffic environment and was used to create statistics of vehicle IDs and their driving routes. Detailed statistics of the vehicle data in each scenario are summarized in Table \ref{table:dataset}. From the table,we can see the Jinan scenario has the fewest intersections, the Hangzhou scenario has the smallest traffic flow, and the synthetic scenario has the highest traffic flow and the most intersections. Intuitively, the higher the traffic flow, the more important the cooperation between intersections. The proposed synthetic scenario is thus designed to test the performance of the proposed method in a complex traffic environment.

\begin{table}[H] \centering
\caption{Data statistics of three traffic datasets}\label{table:dataset}
\begin{tabular}{lccccr}
\hline
\multirow{2}{*}{Dateset} & \multirow{2}{*}{intersections} &\multicolumn{4}{c}{Arrival rate(vehicles/300s)}\\
& &Mean&Std&Max&Min\\
\cline{1-6}
$D_{Jinan}$&12&524.58&98.53&672&256\\
\cline{1-6}
$D_{Hangzhou}$&16&40.45&99.19&333&212\\
\cline{1-6}
$D_{Synthetic}$&16&935.92&17.47&960&896\\
\hline
\end{tabular}
\end{table}

\begin{figure*}[ht]
\centering
\vspace{-0.20cm} 
\subfigtopskip=3pt 
\subfigbottomskip=-2pt
\subfigcapskip=5pt 
\subfigure[Jinan]{
\begin{minipage}{0.31\linewidth}\label{fig:3_4_train}
\centering
\includegraphics[width=5.6cm]{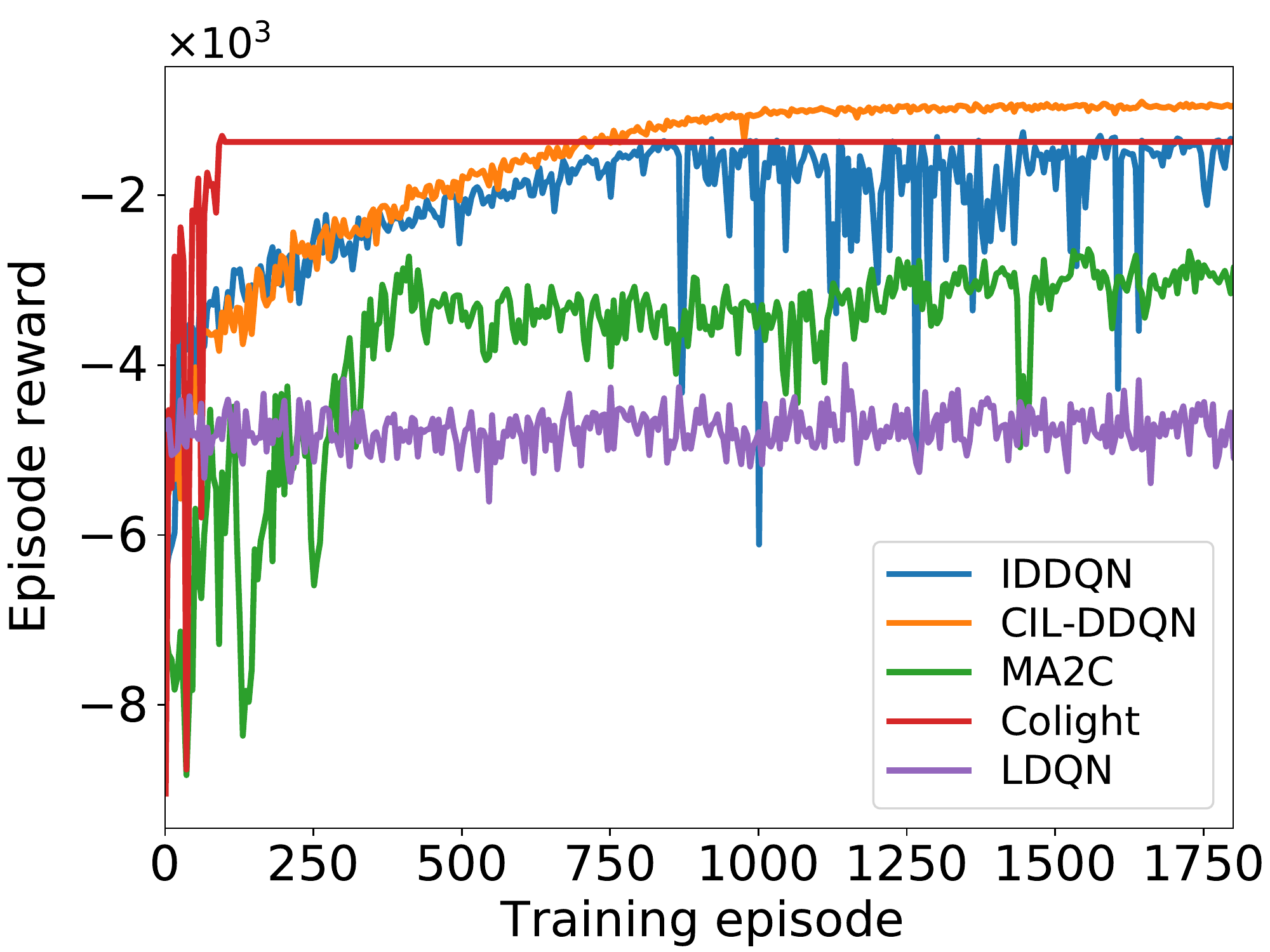}
\end{minipage}
 }
\hfill
\subfigure[Hangzhou]{
\begin{minipage}{0.31\linewidth}\label{fig:4_4_train}
\centering
\includegraphics[width=5.6cm]{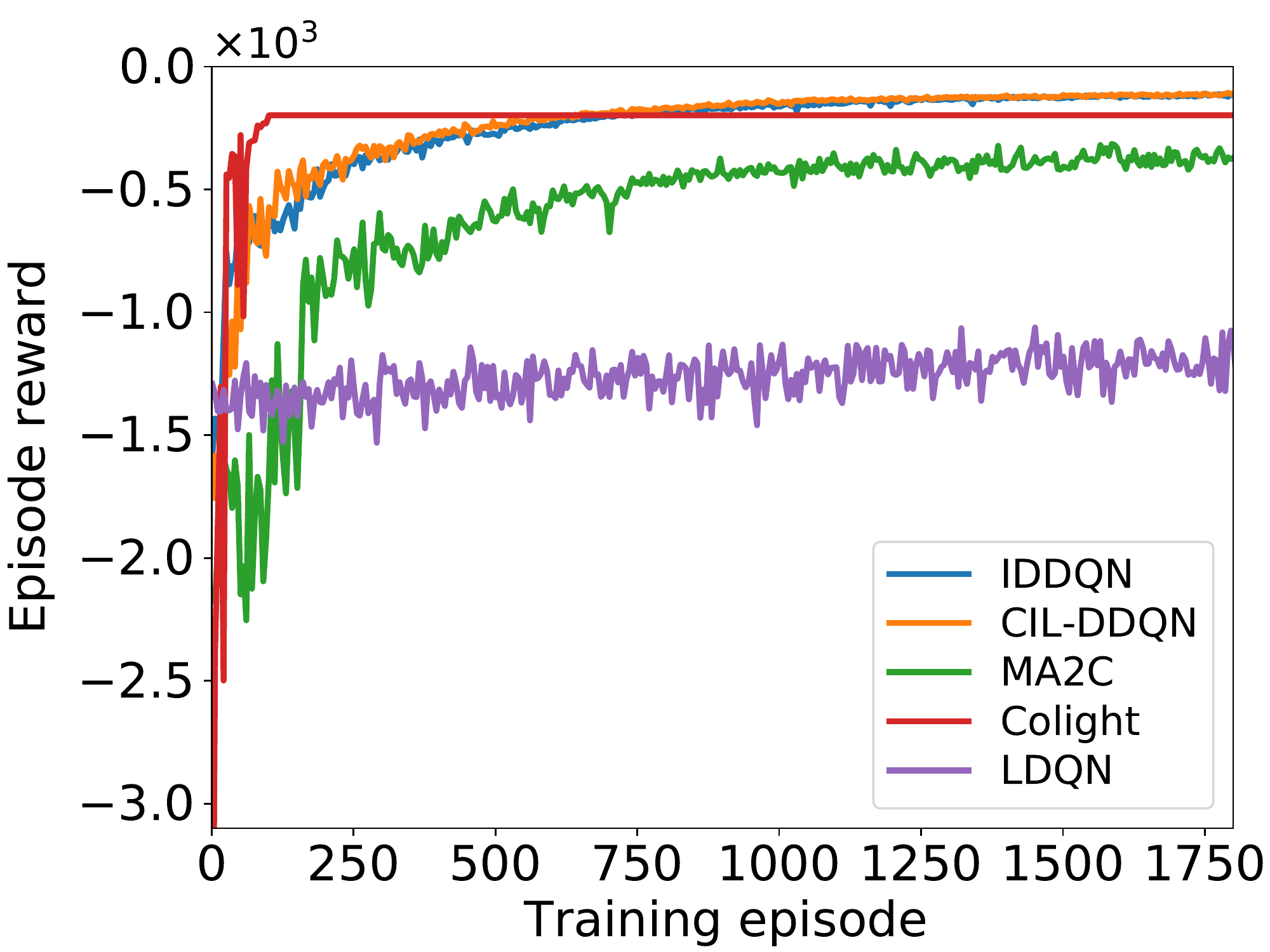}
\end{minipage}%
}
\hfill
\subfigure[synthetic]{
\begin{minipage}{0.31\linewidth} \label{fig:mydata_train}
\centering
\includegraphics[width=5.6cm]{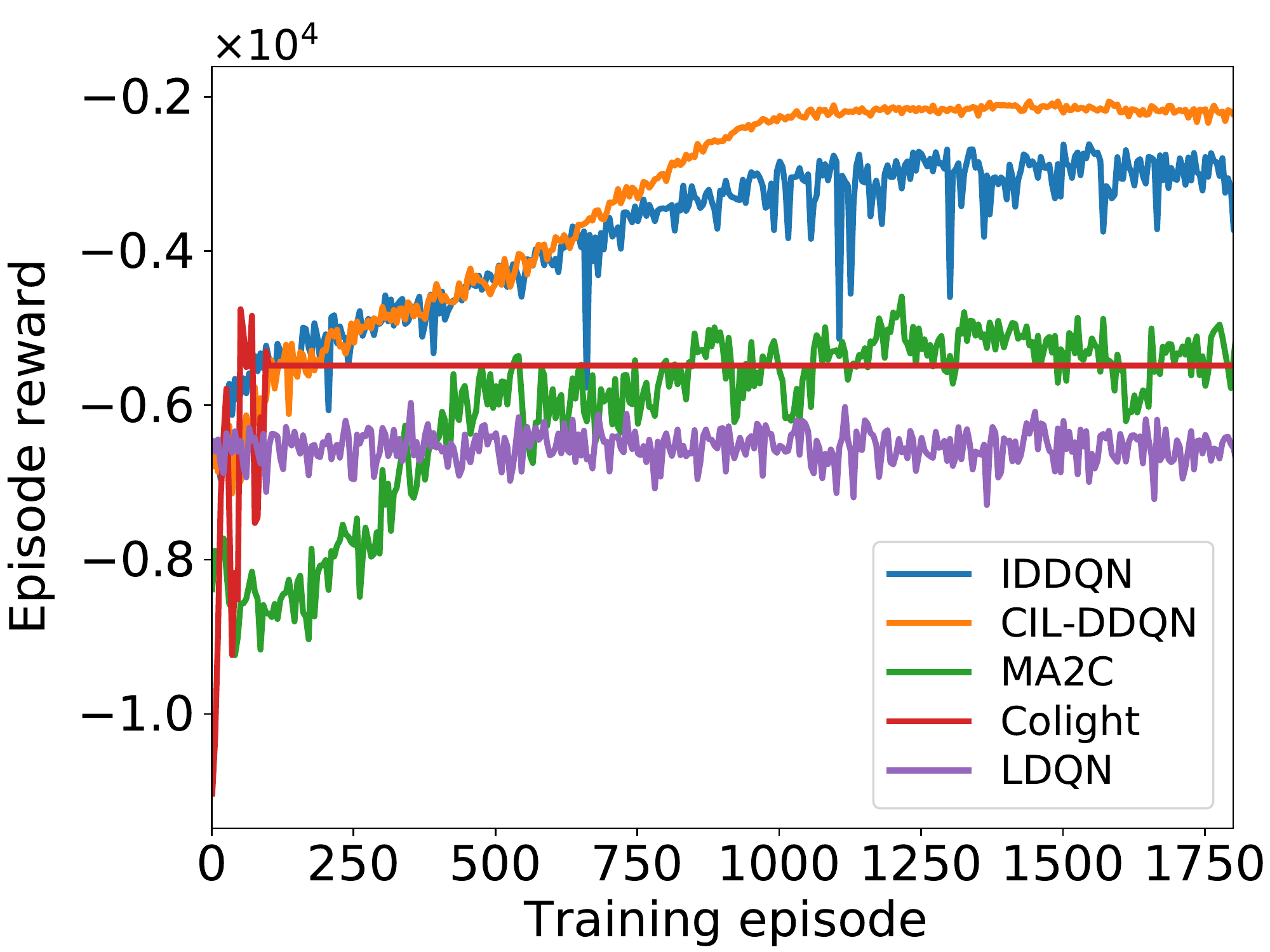}
\end{minipage}%
}
\caption{Average cumulative rewards during all the training episodes in the Jinan (a), Hangzhou (b) and synthetic (c) traffic network.}
\label{fig:3_train}
\end{figure*}

\begin{figure*}[ht]
\centering
\vspace{-0.20cm} 
\subfigtopskip=3pt 
\subfigbottomskip=-2pt
\subfigcapskip=5pt 
\subfigure[Queue length (Jinan)]{
\begin{minipage}{0.31\linewidth} \label{fig:3_4_que}
\centering
\includegraphics[width=5.6cm]{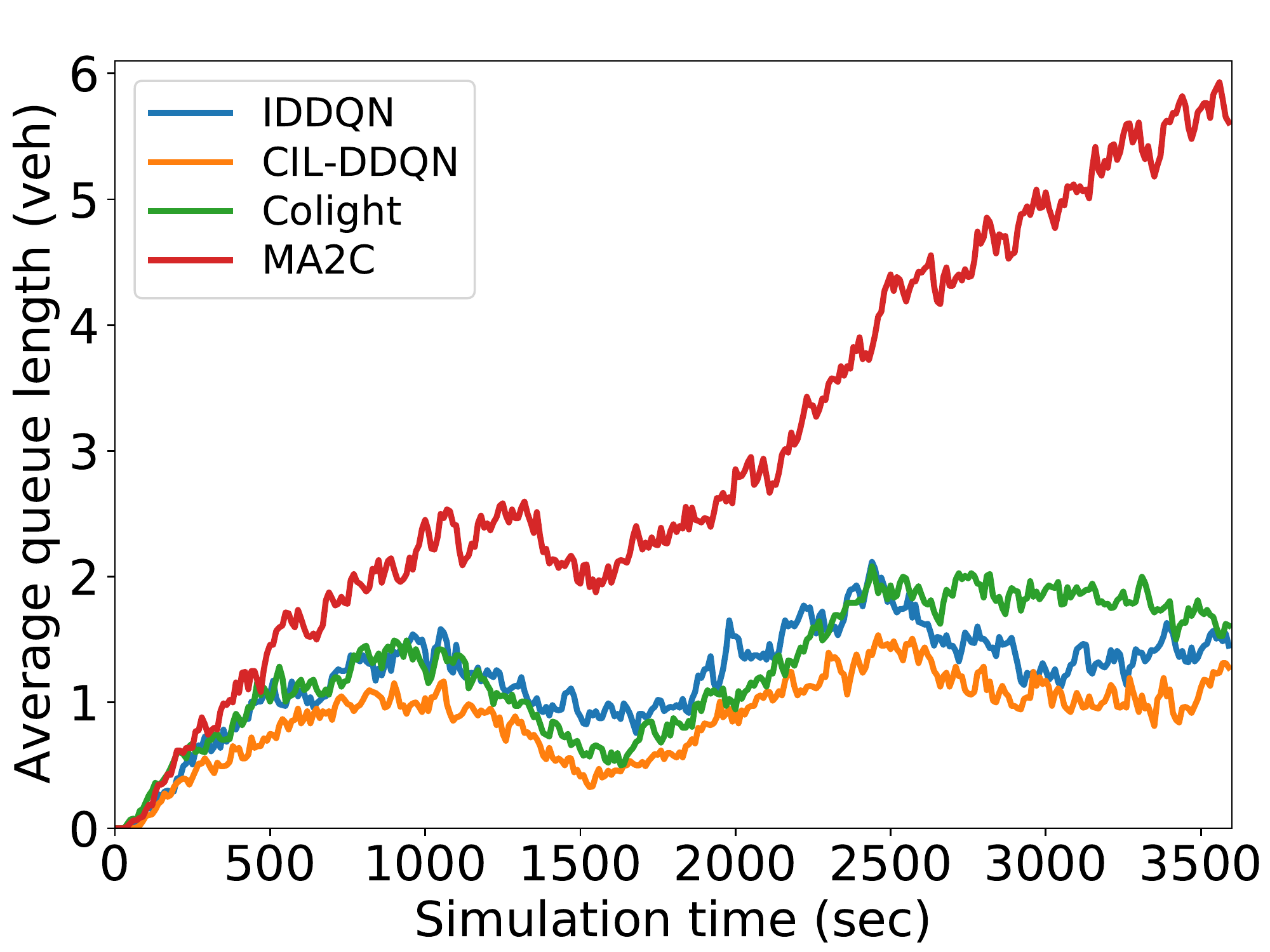}
\end{minipage}
}
\hfill
\subfigure[Queue length (Hangzhou)]{
\begin{minipage}{0.31\linewidth} \label{fig:4_4_que}
\centering
\includegraphics[width=5.6cm]{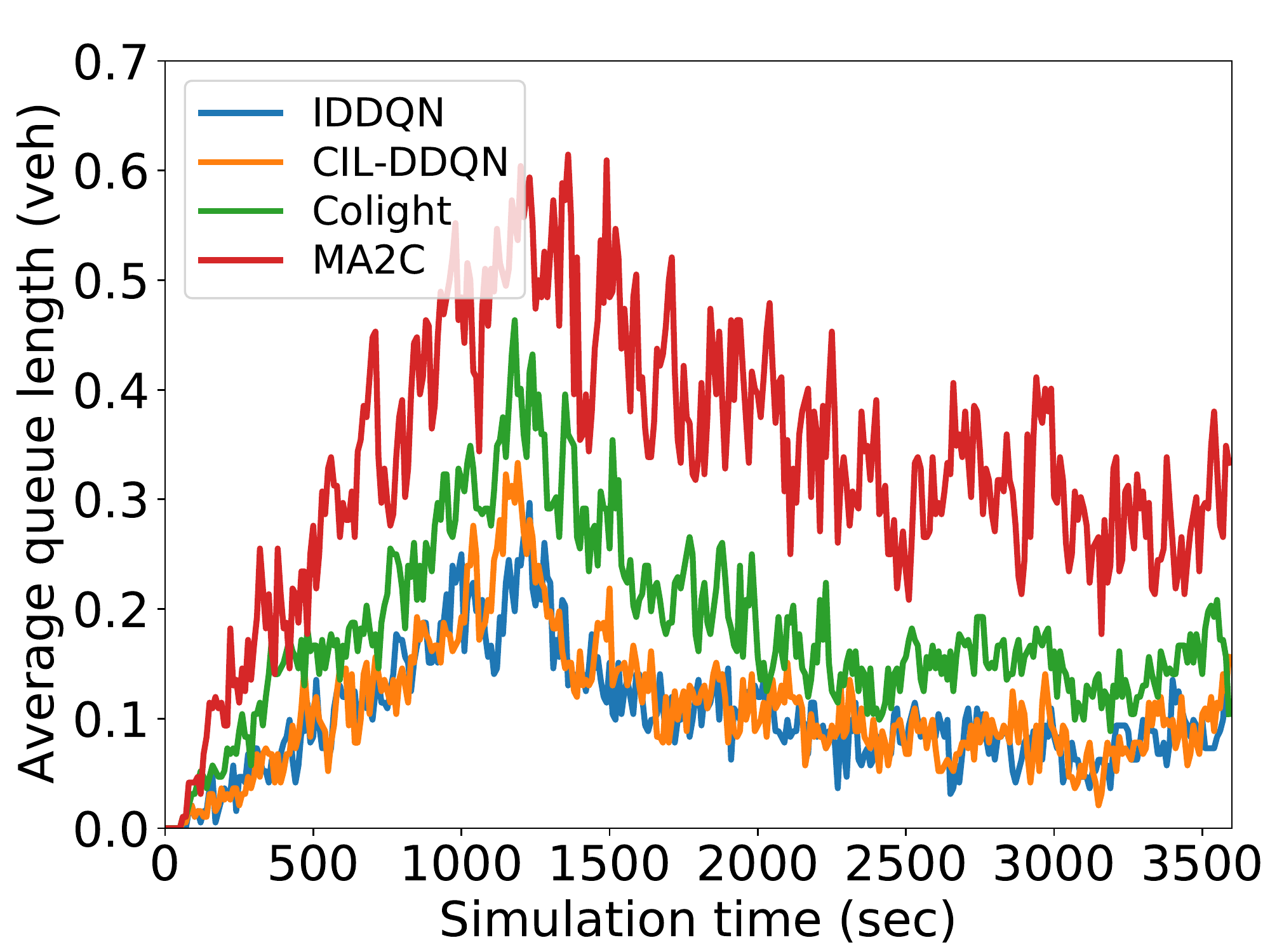}
\end{minipage}%
}
\hfill
\subfigure[Queue length (synthetic)]{
\begin{minipage}{0.31\linewidth} \label{fig:mydata_que}
\centering
\includegraphics[width=5.6cm]{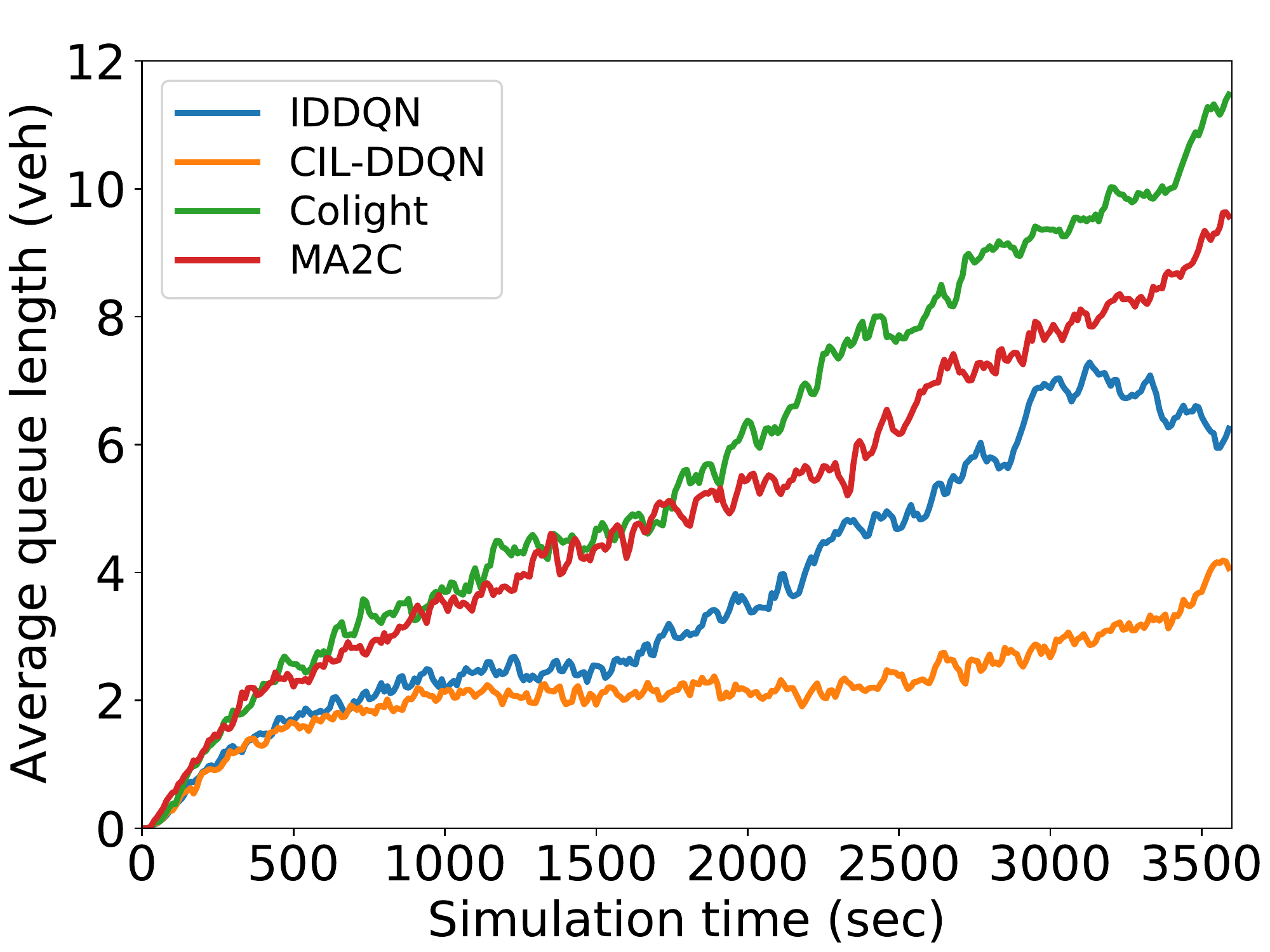}
\end{minipage}%
}
\caption{Average number of waiting vehicles within an hour for each trained RL method in the Jinan (a), Hangzhou (b) and synthetic (c) traffic network.}
\label{fig:4_que}
\end{figure*}

The proposed model aims to minimize traffic congestion across the entire network, which reduces the cumulative delay of all vehicles. Thus, we evaluate the performance of the proposed model using the following three metrics: (1) throughput: the number of vehicles that complete their journey in the road network during an episode, (\ie the number of vehicles arriving at their intended destination); (2) travel time: the average travel time of all vehicles in the road network, which is the most common metric used to evaluate traffic signal control methods in the transportation field; (3) queue length: the average length of queued vehicles in each lane of each intersection during an episode. Specifically, these metrics are defined as follows:
\begin{equation}
\label{eq:metrics}
\begin{split}
{\begin{array}{*{20}{l}}
{{m_{throughout}} =\rm{|}}{{\rm{N}}_v}{\rm{|}}\\
{{m_{trval - time}} = \frac{1}{{|{\rm{N}}_{in}|}}\sum\limits_{v \in {\rm{N}}_{in}} {(T_i^{out} - T_i^{in})} }\\
{{m_{queue - length}} = \frac{1}{{|{\cal N}|}}\sum\limits_{i \in {\cal N}} {\sum\limits_t {\sum\limits_{l \in {L_i}} {\frac{{wait_{\rm{t}}[l]}}{{|{L_i}|}}} } } }
\end{array}}
\end{split}
\end{equation}
where $T_i^{in}$ and $T_i^{out}$ are the arriving and leaving time of vehicle $v$, respectively, ${{\rm{N}}_v}= \{v|0<T_i^{in}<T_i^{out}<3600\}$ is the vehicle set that arrives at their intended destination within time interval $[0,3600]$, ${\rm{N}}_{in}$ is the vehicle set that incoming the traffic network, and $|{\cal N}|$ is the number of intersections.

\subsection{Training setting}
\label{parameter}
Unless otherwise stated, all networks use the same architecture and hyper-parameters. We perform evaluations using a DDQN architecture~\cite{van2016deep} as a basis for the algorithm. The Q networks of LDQN, IDDQN, and CIL-DDQN consist of 2 fully connected layers with 200 neurons, and an output neuron for each action. To compare results fairly, we ran the Fixedtime, SOTL, MA2C and Colight algorithms with the source code\footnote{MA2C: \url{https://github.com/cts198859/deeprl-signal-control}; Fixedtime, SOTL and Colight: \url{https://github.com/wingsweihua/colight}} released by the authors and used identical parameters. Other parameters are summarized in Table \ref{parametrer}.

\begin{table}[ht]
\centering \caption{Parameter setting}
\label{parametrer}
\begin{tabular}{l|l|l}
\hline
Component                                         & Hyper-parameter                                   &     Setting                       \\ \hline \hline
\multirow{6}{*}{DDQN-optimization}                & Learning rate $\alpha$                            & 0.001                             \\ \cline{2-3}
                                                  & Discount rate $\gamma$                            & 0.9                               \\ \cline{2-3}
                                                  & ERM size                                          & 200000                            \\ \cline{2-3}
                                                  & Soft update weighting factor $\tau$                     & 0.001                             \\ \cline{2-3}
                                                  & Network optimizer                                 & Adam
                                                           \\ \cline{2-3}
                                                  & Activation function
                                                  & Relu
                                                  \\ \hline \hline
\multirow{2}{*}{$\epsilon$-greedy exploration} & Initial and final $\epsilon$ values            & 0.8 and 0.001                       \\ \cline{2-3}
                                                  & Decay rate $d_\epsilon$                        & 0.8/360000
                                                  \\ \hline \hline
\multirow{4}{*}{LDQN}                             & Initial temperature value                         & 1                                 \\ \cline{2-3}
                                                  & Temperature decay value $\mu$                     & 2                                 \\ \cline{2-3}
                                                  & Leniency modification value $K$                   & 2                                 \\ \cline{2-3}
                                                  & TDS exponent decay rate d                         & 0.95                              \\ \hline \hline
\multirow{3}{*}{CIL-DDQN}                         &Initial and final $l$ values                       &0.5 and 0
\\ \cline{2-3}
                                                  &Leniency decay rate $d_l$                          &0.5/800000
\\ \cline{2-3}
                                                  &Importance decay rate $d_e$                        &0.995

\\\hline \hline
\end{tabular}
\end{table}

\begin{table*}[ht]
\centering
\caption{Execution performance comparison over trained MARL policies. Best values are in bold.}\label{table:Jinnan Performance}
\begin{tabular}{l|l|ccccccc}
\hline
\multicolumn{2}{c|}{Metrics} & SOTL & Fixedtime & Colight & MA2C & LDQN & IDDQN & CIL-DDQN (ours)\\
\hline
\multirow{3}{*}{Jinan} &Travel time(s)&1459.08&811.92&294.36&432.66&569.27&331.96&\textbf{255.10}\\
                       &Queue length(veh)&13.24&7.80&1.27&3.00&4.23&1.76&\textbf{0.84}\\
                       &Throughput(veh/h)&1508&3475&5725&5117&4410&5500&\textbf{5864}\\
\hline
\multirow{3}{*}{Hangzhou}
                       &Travel time(s)&1209.33&705.31&284.07&321.41&514.68&265.74&\textbf{263.69}\\
                       &Queue length(veh)&4.19&2.22&0.18&0.34&1.10&0.11&\textbf{0.10}\\
                       &Throughput(veh/h)&1076&2000&2782&2700&2374&2795&\textbf{2795}\\
\hline
\multirow{3}{*}{Synthetic}
                       &Travel time(s)&1461.85&869.88&563.49&487.93&711.51&360.77&\textbf{250.37}\\
                       &Queue length(veh)&11.04&9.75&5.80&5.04&6.09&3.68&\textbf{2.10}\\
                       &Throughput(veh/h)&2306&5022&6908&7630&5463&9107&\textbf{10251}\\
\hline
\end{tabular}
\end{table*}

\subsection{Performance comparison with state-of-the-art methods}

Figure \ref{fig:3_train} shows the training curves of all MARL methods mentioned above in the Jinan (a), Hangzhou (b), and synthetic (c) traffic networks. The lines in each picture of Figure \ref{fig:3_train} indicate the average cumulative reward during all the training episodes of each method. Figure \ref{fig:3_train} shows that CIL-DDQN outperforms all other methods with regard to learning effectiveness in all three scenarios. Among the other methods, the most advanced RL traffic method Colight achieves the fastest convergence and produces good results in the two real-world scenarios but does not perform well in the synthetic scenario. Besides, IDDQN yields almost the same results as CIL-DDQN in the Hangzhou scenario and outperforms the other methods except CIL-DDQN in the other two scenarios. LDQN performs the worst among examined methods. In the synthetic and Jinan scenarios, the higher traffic flow increases the coordination demand between intersections and thus increases the training difficulty. CIL-DDQN performs markedly better than the other RL algorithms. In the Hangzhou scenario, the lower traffic flow makes the intersections almost act independently from each other; thus, the single-agent algorithm IDDQN achieves the best performance of all the algorithms except for CIL-DDQN. In the other two scenarios, CIL-DDQN achieves better stability and convergence compared to IDDQN. These results show that CIL-DDQN can effectively regulate the relationship between various intersections with different levels of traffic.


Table \ref{table:Jinnan Performance} shows the results of the execution performance comparison with trained MARL policies. We also show the average number of waiting vehicles within an hour for each trained RL method in the Jinan (a), Hangzhou (b), and synthetic (c) traffic network (see Figure \ref{fig:4_que}) for completeness. Because the LDQN algorithm failed to converge, we did not include it in the figure. Table \ref{table:Jinnan Performance} and Figure \ref{fig:4_que} show that the proposed method outperforms other methods in all metrics.


\subsection{Validation of PO-WCTM model}
\begin{figure}[H]
\centering
\setlength{\abovecaptionskip}{0.cm}
\includegraphics[width=5.0cm]{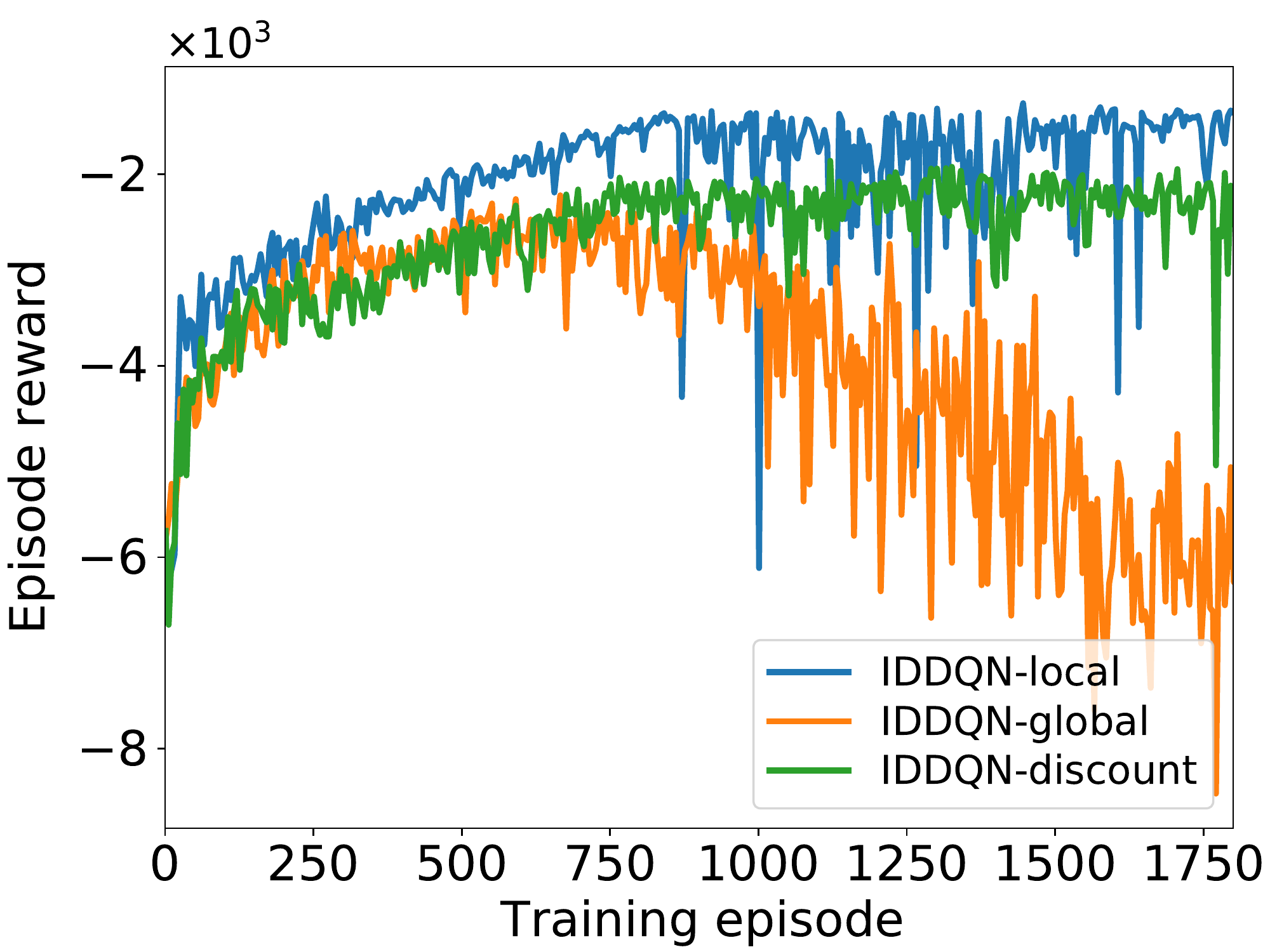}
\caption{Average cumulative rewards during all the training episodes of IDDQN under different reward settings in the Jinan scenario}
\label{fig:3_4IDDQN} 
\end{figure}

\begin{figure}[H]
\centering
\setlength{\abovecaptionskip}{0.cm}
\includegraphics[width=5.0cm]{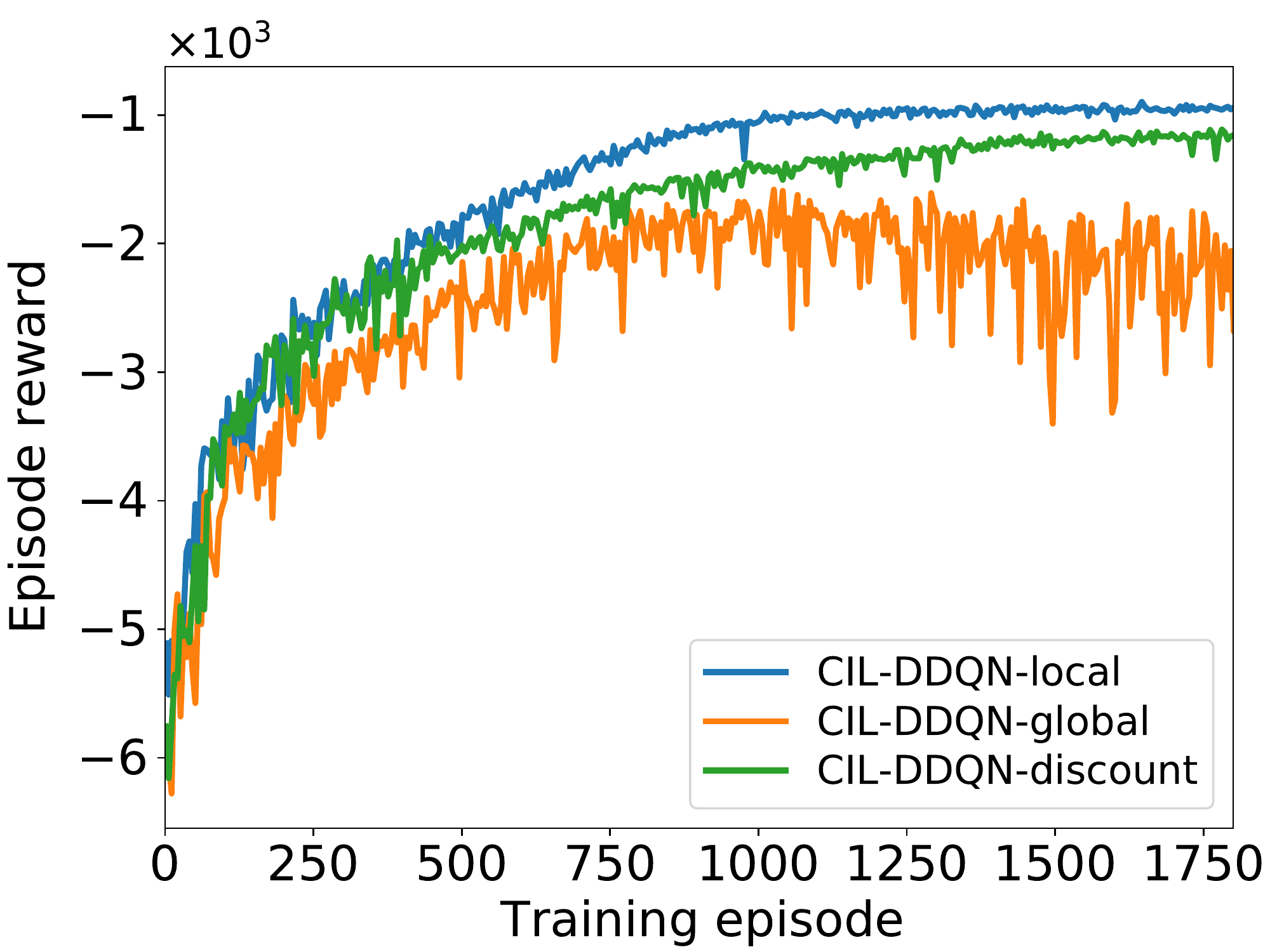}
\caption{Average cumulative rewards during all the training episodes of CIL-DDQN under different reward settings in the Jinan scenario}
\label{fig:3_4CIL-DDQN} 
\end{figure}

To verify the rationality of the PO-WCTM model, we tested the proposed PO-WCTM model with three reward settings (local, global, and discount) with CIL-DDQN and IDDQN in all three traffic scenarios. The local reward setting is the average length of waiting vehicles on each incoming lane of an agent as well as its neighbors. The global reward setting is the average length of waiting vehicles of all agents. The discount reward setting is the average length of waiting vehicles of all agents weighted by distance between agents, as in MA2C~\cite{chu2020multi-agent}. Table \ref{table:Model verification} shows the final results of all reward settings with the metric $m_{trval-time}$ (Eq.\ref{eq:metrics}). Figure \ref{fig:3_4IDDQN} and Figure \ref{fig:3_4CIL-DDQN} show the training curves of the two algorithms in the Jinan scenario with different reward settings. Experimental results show that the coordination of local intersections can promote the coordination of all intersections across the whole road network, which demonstrates the validity of PO-WCTM.
\begin{table}[H] \centering
\caption{Performance on all datasets with travel time}\label{table:Model verification}
\begin{tabular}{l|ccc}
\hline
&Jinan&Hangzhou&Synthetic\\
\hline
IDDQN - local reward&331.96&265.74&360.77\\
\hline
CIL-DDQN -local reward&\textbf{255.10}&\textbf{263.69}&\textbf{250.37}\\
\hline
IDDQN - global reward&639.44&314.67&749.13\\
\hline
CIL-DDQN - global reward&367.62&294.66&290.90\\\hline
IDDQN - discount reward&349.32&286.13&442.36\\\hline
CIL-DDQN - discount reward&275.30&276.21&259.07\\
\hline
\end{tabular}
\end{table}

\vspace{-0.5cm}

\subsection{Parameter study}

\begin{figure}[H]
\centering
\setlength{\abovecaptionskip}{0.cm}
\includegraphics[width=5.0cm]{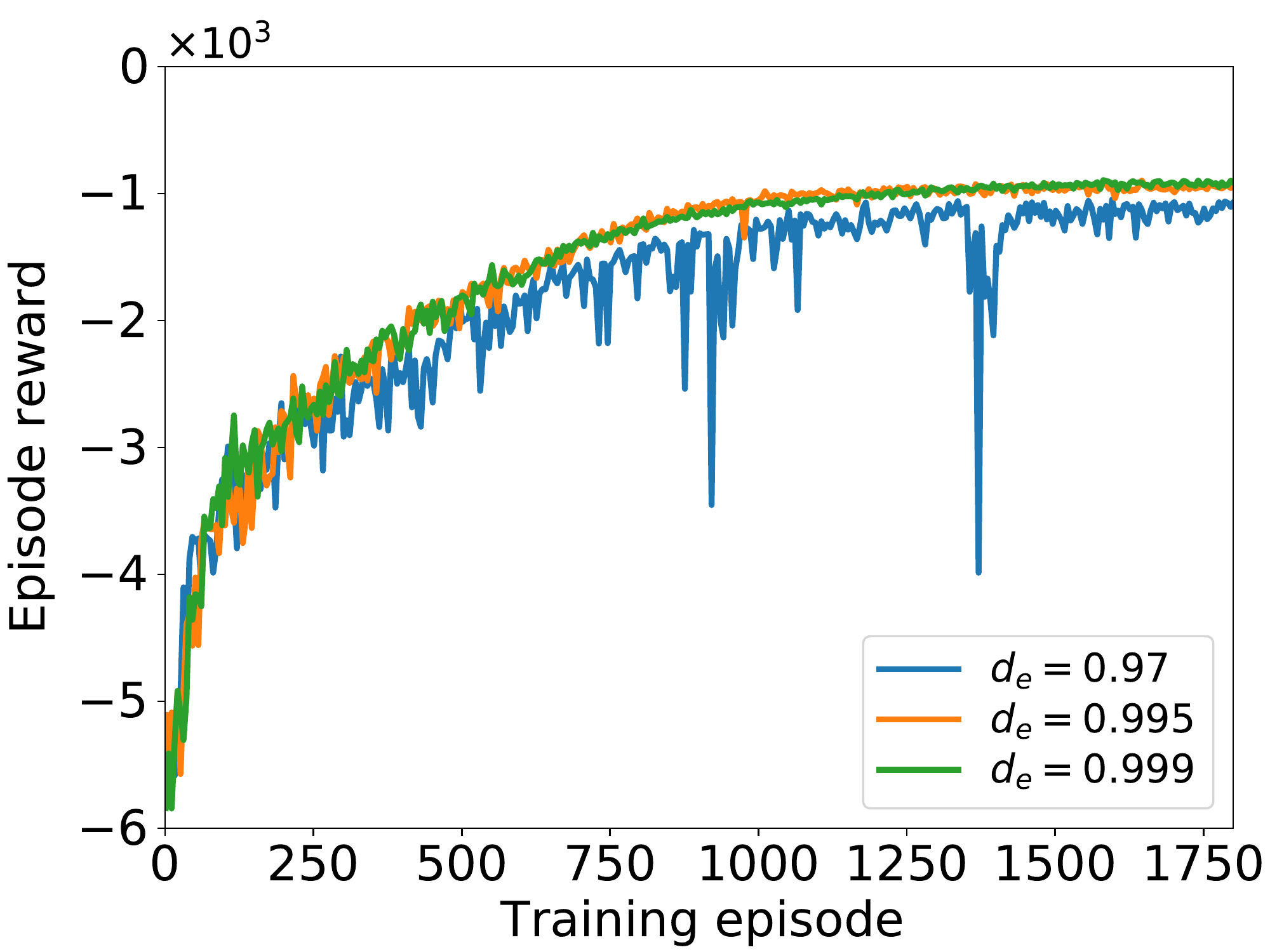}
\caption{Average cumulative rewards during all the training episodes of CIL-DDQN under different settings of $d_e$ in the Jinan scenario}
\label{fig:de} 
\end{figure}

\begin{figure}[H]
\centering
\setlength{\abovecaptionskip}{0.cm}
\includegraphics[width=5.0cm]{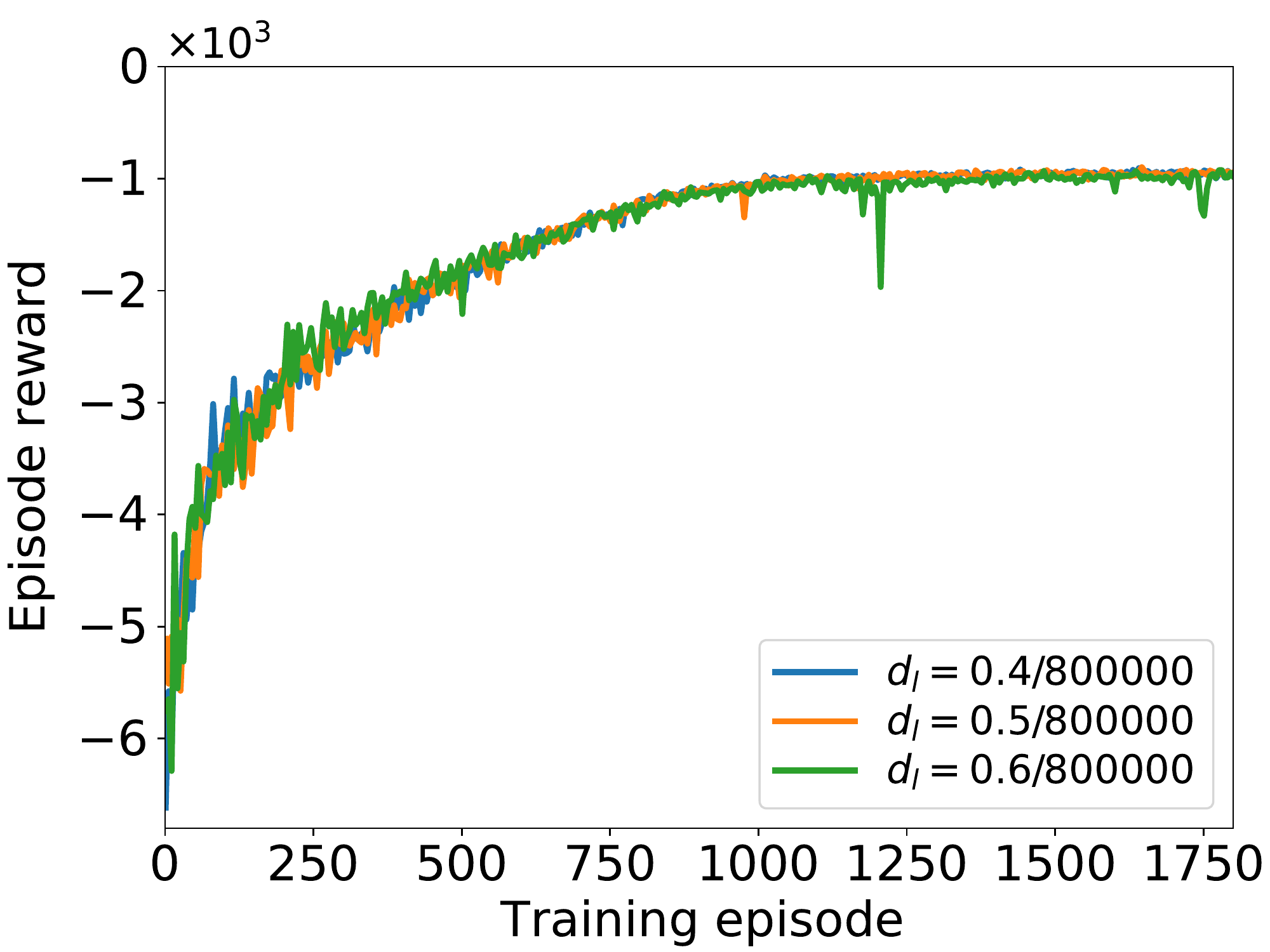}
\caption{Average cumulative rewards during all the training episodes of CIL-DDQN under different settings of $d_l$ in the Jinan scenario}
\label{fig:dl} 
\end{figure}

In this section, we analyze the robustness of the two key parameters in CIL-DDQN, \ie, the importance decay rate $d_e$ and the leniency decay rate $d_l$, corresponding to the two mechanisms of CIL-DDQN. Figure \ref{fig:de} and \ref{fig:dl} show the learning performance of the proposed method with different settings of $d_e$ and $d_l$ in the Jinan scenario. Experimental results show that, except for some slight oscillations in $d_e=0.97$, the performances of the proposed method achieve the optimal return almost under all the different parameter settings, which indicates that our method is robust in the two parameters.
\vspace{-0.5cm}

\section{Conclusion}\label{sec:con}
To address the problem of global cooperation in the domain of traffic signal control, this paper proposed PO-WCTM, a partially observable weak cooperative traffic model that considers the network relationships of traffic intersections. Based on PO-WCTM, we proposed CIL-DDQN, an multiagent cooperative method extended by LDQN and HDQN, to learn the cooperative control policies to optimize the overall throughput of cars through multiple intersections. We evaluated the proposed method in one synthetic and two real-world traffic scenarios, and compared its performance with the other state-of-the-art methods. Experimental results verified the effectiveness of the proposed model and algorithm, where each agent computes its policy locally with information about its neighboring intersections. Generally, the proposed algorithm can scale up to large problems with large numbers of intersections if necessary.

In the future, we plan to design algorithms that can optimally split the traffic network into regions and test the proposed algorithm with data from real traffic flow in very large traffic networks (\eg, a whole city). Another important direction of future research is defining the state design of agents in traffic signal control problems, by analyzing the weights of information about neighbors and designing a framework of neural networks to extract effective traffic characteristics.


%

\section*{Acknowledgment}

The work is supported by the National Natural Science Foundation of China (Grant
Nos.:61906027, 61906135), China Postdoctoral Science Foundation Funded Project (Grant No.:2019M661080).

\ifCLASSOPTIONcaptionsoff
  \newpage
\fi


\bibliographystyle{IEEEtran}
\bibliography{bib/IEEEexample}

\begin{thebibliography}{10}
\providecommand{\url}[1]{#1}
\csname url@samestyle\endcsname
\providecommand{\newblock}{\relax}
\providecommand{\bibinfo}[2]{#2}
\providecommand{\BIBentrySTDinterwordspacing}{\spaceskip=0pt\relax}
\providecommand{\BIBentryALTinterwordstretchfactor}{4}
\providecommand{\BIBentryALTinterwordspacing}{\spaceskip=\fontdimen2\font plus
\BIBentryALTinterwordstretchfactor\fontdimen3\font minus
  \fontdimen4\font\relax}
\providecommand{\BIBforeignlanguage}[2]{{%
\expandafter\ifx\csname l@#1\endcsname\relax
\typeout{** WARNING: IEEEtran.bst: No hyphenation pattern has been}%
\typeout{** loaded for the language `#1'. Using the pattern for}%
\typeout{** the default language instead.}%
\else
\language=\csname l@#1\endcsname
\fi
#2}}
\providecommand{\BIBdecl}{\relax}
\BIBdecl

\bibitem{liang2019a}
X.~Liang, X.~Du, G.~Wang, and Z.~Han, ``A deep reinforcement learning network
  for traffic light cycle control,'' \emph{IEEE Transactions on Vehicular
  Technology}, vol.~68, no.~2, pp. 1243--1253, 2019.

\bibitem{gartner1990multiband--a}
N.~H. Gartner, S.~F. Assmann, F.~Lasaga, and D.~L. Hous, ``Multiband--a
  variable-bandwidth arterial progression scheme,'' \emph{Transportation
  Research Record}, no. 1287, 1990.

\bibitem{cools2006self-organizing}
S.~Cools, C.~Gershenson, and B.~Dhooghe, ``Self-organizing traffic lights: A
  realistic simulation,'' \emph{arXiv: Adaptation and Self-Organizing Systems},
  2006.

\bibitem{luk1984two}
J.~Y. Luk, ``Two traffic-responsive area traffic control methods: Scat and
  scoot,'' \emph{Traffic engineering and control}, vol.~25, no.~1, pp. 14--22,
  1984.

\bibitem{hunt1982the}
P.~B. Hunt, D.~I. Robertson, R.~D. Bretherton, and M.~C. Royle, ``The scoot
  on-line traffic signal optimisation technique,'' \emph{Traffic engineering
  and control}, vol.~23, no.~4, 1982.

\bibitem{nishi2018traffic}
T.~Nishi, K.~Otaki, K.~Hayakawa, and T.~Yoshimura, ``Traffic signal control
  based on reinforcement learning with graph convolutional neural nets,'' pp.
  877--883, 2018.

\bibitem{wei2019colight}
H.~Wei, N.~Xu, H.~Zhang, G.~Zheng, X.~Zang, C.~Chen, W.~Zhang, Y.~Zhu, K.~Xu,
  and Z.~Li, ``Colight: Learning network-level cooperation for traffic signal
  control,'' pp. 1913--1922, 2019.

\bibitem{zhu2015a}
F.~Zhu, H.~M.~A. Aziz, X.~Qian, and S.~V. Ukkusuri, ``A junction-tree based
  learning algorithm to optimize network wide traffic control: A coordinated
  multi-agent framework,'' \emph{Transportation Research Part C-emerging
  Technologies}, vol.~58, pp. 487--501, 2015.

\bibitem{bloembergen2015evolutionary}
D.~Bloembergen, K.~Tuyls, D.~Hennes, and M.~Kaisers, ``Evolutionary dynamics of
  multi-agent learning: a survey,'' \emph{Journal of Artificial Intelligence
  Research}, vol.~53, no.~1, pp. 659--697, 2015.

\bibitem{matignon2012review}
L.~Matignon, G.~J. Laurent, and N.~L. Fortpiat, ``Review: independent
  reinforcement learners in cooperative markov games: A survey regarding
  coordination problems,'' \emph{Knowledge Engineering Review}, vol.~27, no.~1,
  pp. 1--31, 2012.

\bibitem{sunehag2017value-decomposition}
P.~Sunehag, G.~Lever, A.~Gruslys, W.~M. Czarnecki, V.~Zambaldi, M.~Jaderberg,
  M.~Lanctot, N.~Sonnerat, J.~Z. Leibo, K.~Tuyls \emph{et~al.},
  ``Value-decomposition networks for cooperative multi-agent learning.''
  \emph{arXiv: Artificial Intelligence}, 2017.

\bibitem{lowe2017multi-agent}
R.~Lowe, Y.~Wu, A.~Tamar, J.~Harb, P.~Abbeel, and I.~Mordatch, ``Multi-agent
  actor-critic for mixed cooperative-competitive environments,'' pp.
  6379--6390, 2017.

\bibitem{son2019qtran}
K.~Son, D.~Kim, W.~J. Kang, D.~E. Hostallero, and Y.~Yi, ``Qtran: Learning to
  factorize with transformation for cooperative multi-agent reinforcement
  learning,'' pp. 5887--5896, 2019.

\bibitem{wei2016lenient}
E.~Wei and S.~Luke, ``Lenient learning in independent-learner stochastic
  cooperative games,'' \emph{Journal of Machine Learning Research}, vol.~17,
  no.~1, pp. 2914--2955, 2016.

\bibitem{omidshafiei2017deep}
S.~Omidshafiei, J.~Pazis, C.~Amato, J.~P. How, and J.~Vian, ``Deep
  decentralized multi-task multi-agent reinforcement learning under partial
  observability,'' in \emph{Proceedings of the 34th International Conference on
  Machine Learning-Volume 70}.\hskip 1em plus 0.5em minus 0.4em\relax JMLR.
  org, 2017, pp. 2681--2690.

\bibitem{palmer2018lenient}
G.~Palmer, K.~Tuyls, D.~Bloembergen, and R.~Savani, ``Lenient multi-agent deep
  reinforcement learning,'' pp. 443--451, 2018.

\bibitem{HaoWHY19}
X.~Hao, W.~Wang, J.~Hao, and Y.~Yang, ``Independent generative adversarial
  self-imitation learning in cooperative multiagent systems,'' in
  \emph{Proceedings of the 18th International Conference on Autonomous Agents
  and MultiAgent Systems, {AAMAS}2019}, 2019.

\bibitem{palmer2019negative}
G.~Palmer, R.~Savani, and K.~Tuyls, ``Negative update intervals in deep
  multi-agent reinforcement learning,'' in \emph{Proceedings of the 18th
  International Conference on Autonomous Agents and MultiAgent Systems}, 2019.

\bibitem{van2016deep}
H.~Van~Hasselt, A.~Guez, and D.~Silver, ``Deep reinforcement learning with
  double q-learning,'' pp. 2094--2100, 2016.

\bibitem{chu2020multi-agent}
T.~Chu, J.~Wang, L.~Codeca, and Z.~Li, ``Multi-agent deep reinforcement
  learning for large-scale traffic signal control,'' \emph{IEEE Transactions on
  Intelligent Transportation Systems}, vol.~21, no.~3, pp. 1086--1095, 2020.

\bibitem{Wiering2004Intelligent}
J.~V. Wiering, Jelle Van~Veenen and A.~Koopman, ``Intelligent traffic light
  control,'' \emph{Institute of Information and Computing Sciences.Utrecht
  University}, 2004.

\bibitem{li2016traffic}
L.~Li, Y.~Lv, and F.~Wang, ``Traffic signal timing via deep reinforcement
  learning,'' \emph{IEEE/CAA Journal of Automatica Sinica}, vol.~3, no.~3, pp.
  247--254, 2016.

\bibitem{Mnih2015Human}
V.~Mnih, K.~Kavukcuoglu, D.~Silver, A.~A. Rusu, J.~Veness, M.~G. Bellemare,
  A.~Graves, M.~Riedmiller, A.~K. Fidjeland, and G.~Ostrovski, ``Human-level
  control through deep reinforcement learning.'' \emph{Nature}, vol. 518, no.
  7540, p. 529, 2015.

\bibitem{choe2018deep}
C.~Choe, S.~Baek, B.~Woon, and S.~Kong, ``Deep q learning with lstm for traffic
  light control,'' pp. 331--336, 2018.

\bibitem{wang2016dueling}
Z.~Wang, T.~Schaul, M.~Hessel, H.~Hasselt, M.~Lanctot, and N.~Freitas,
  ``Dueling network architectures for deep reinforcement learning,'' in
  \emph{International conference on machine learning}.\hskip 1em plus 0.5em
  minus 0.4em\relax PMLR, 2016, pp. 1995--2003.

\bibitem{Matignon2012}
L.~Matignon, G.~j. Laurent, and N.~Le~fort piat, ``Review: Independent
  reinforcement learners in cooperative markov games: A survey regarding
  coordination problems,'' \emph{Knowl. Eng. Rev.}, vol.~27, no.~1, pp. 1--31,
  2012.

\bibitem{claus1998the}
C.~Claus and C.~Boutilier, ``The dynamics of reinforcement learning in
  cooperative multiagent systems,'' pp. 746--752, 1998.

\bibitem{zheng2019diagnosing}
G.~Zheng, X.~Zang, N.~Xu, H.~Wei, Z.~Yu, V.~V. Gayah, K.~Xu, and Z.~Li,
  ``Diagnosing reinforcement learning for traffic signal control.''
  \emph{arXiv: Learning}, 2019.

\bibitem{chu2019multi}
T.~Chu, S.~Chinchali, and S.~Katti, ``Multi-agent reinforcement learning for
  networked system control,'' in \emph{International Conference on Learning
  Representations}, 2019.

\bibitem{rashid2018qmix}
T.~Rashid, M.~Samvelyan, C.~S. Witt, G.~Farquhar, J.~Foerster, and S.~Whiteson,
  ``Qmix: Monotonic value function factorisation for deep multi-agent
  reinforcement learning,'' in \emph{International Conference on Machine
  Learning}, 2018, pp. 4292--4301.

\bibitem{tan1997multi-agent}
M.~Tan, ``Multi-agent reinforcement learning: independent vs. cooperative
  agents,'' pp. 487--494, 1997.

\bibitem{zhang2019cityflow}
H.~Zhang, S.~Feng, C.~Liu, Y.~Ding, Y.~Zhu, Z.~Zhou, W.~Zhang, Y.~Yu, H.~Jin,
  and Z.~Li, ``Cityflow: A multi-agent reinforcement learning environment for
  large scale city traffic scenario,'' pp. 3620--3624, 2019.

\end{thebibliography}




%

\begin{IEEEbiography}[{\includegraphics[width=1in,height=1.25in,clip,keepaspectratio]{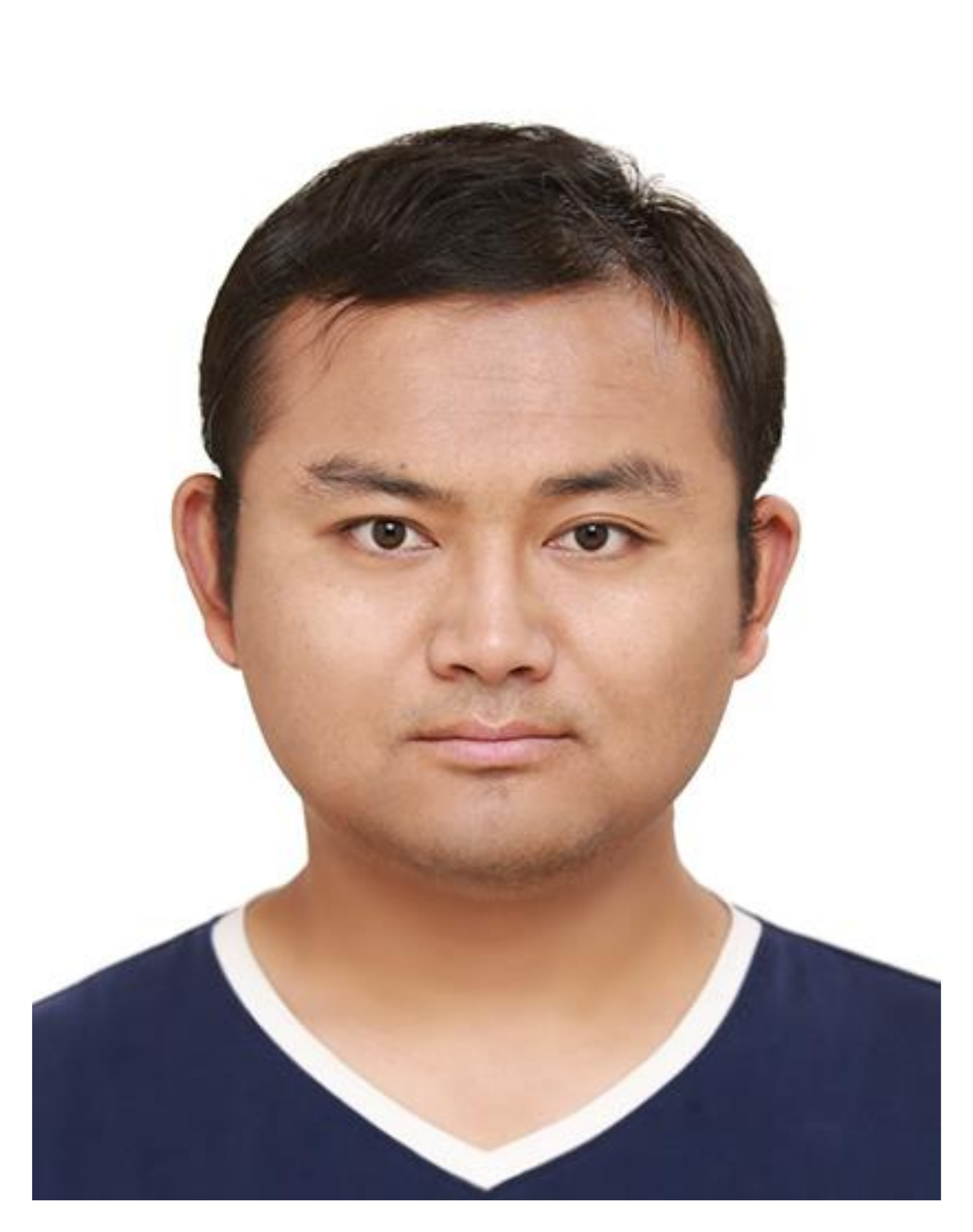}}]
{Chengwei Zhang} is a lecturer of the College of Information Science and Technology at the Dalian Maritime University. He received the Ph.D. in Computer software and theory from Tianjin University in 2018 and B.S. degree in pure and applied mathematics from Northwestern Polytechnical University in 2011. His current research interests include Artificial intelligence, Multi-agent systems and Reinforcement Learning.
\end{IEEEbiography}

\begin{IEEEbiography}[{\includegraphics[width=1in,height=1.25in,clip,keepaspectratio]{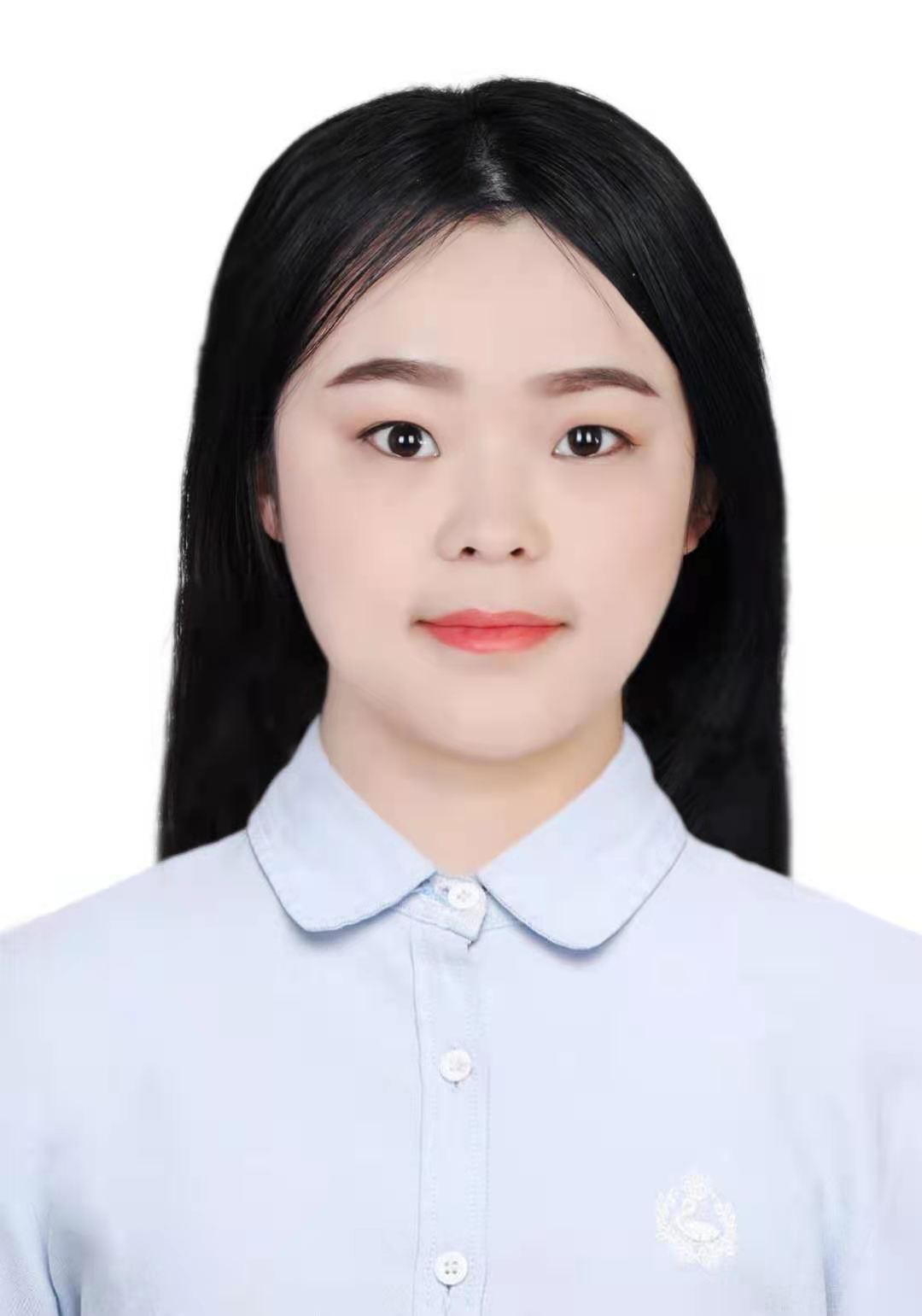}}]
{Shan Jin} received the B.S. degree in Internet of Things engineering from North University of China in 2018. She is currently pursuing the M.Sc. degree in computer science and technology with the Dalian Maritime University. Her current research interests include multi-agent reinforcement learning and traffic signal control.
\end{IEEEbiography}

\begin{IEEEbiography}[{\includegraphics[width=1in,height=1.25in,clip,keepaspectratio]{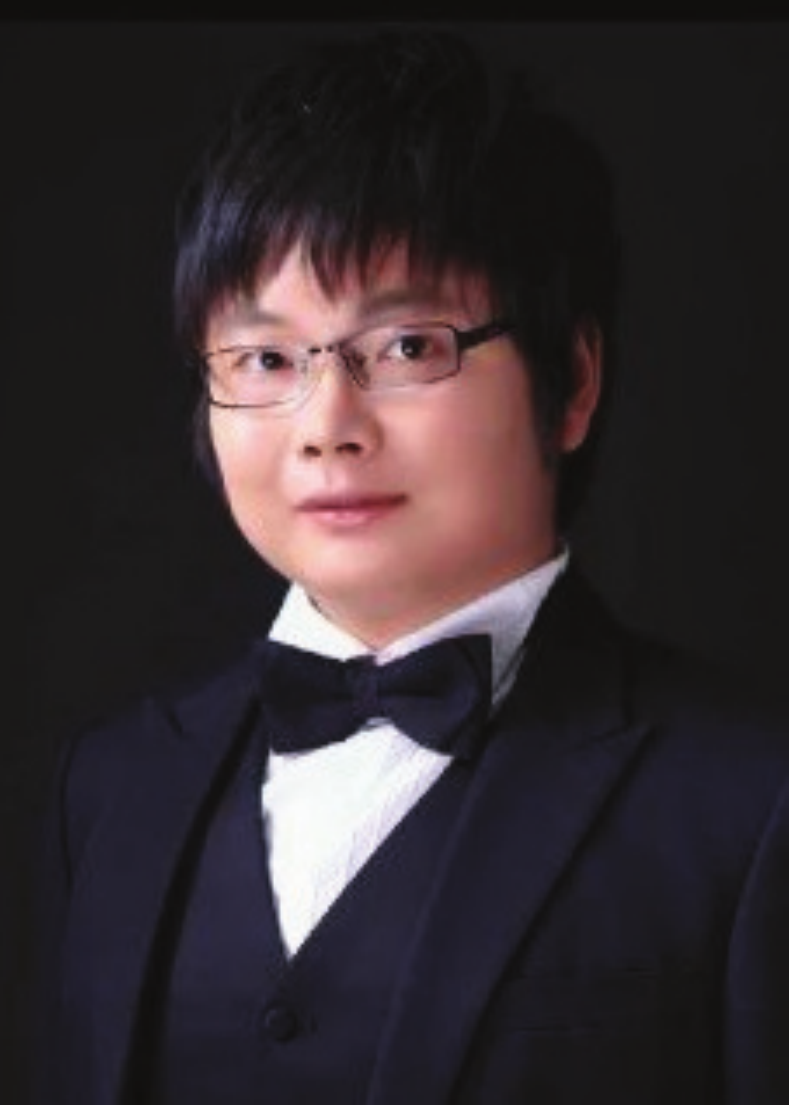}}]
{Wanli Xue} received the B.S. degree in pure and applied mathematics from Tianjin Polytechnic University in 2009 and Ph.D. in Technology of Computer Application from Tianjin University in 2018. Currently, he is a lecturer of the School of Computer Science and Engineering, Tianjin University of Technology. His research interests include visual tracking, machine learning. Corresponding author.
\end{IEEEbiography}


\begin{IEEEbiography}[{\includegraphics[width=1in,height=1.25in,clip,keepaspectratio]{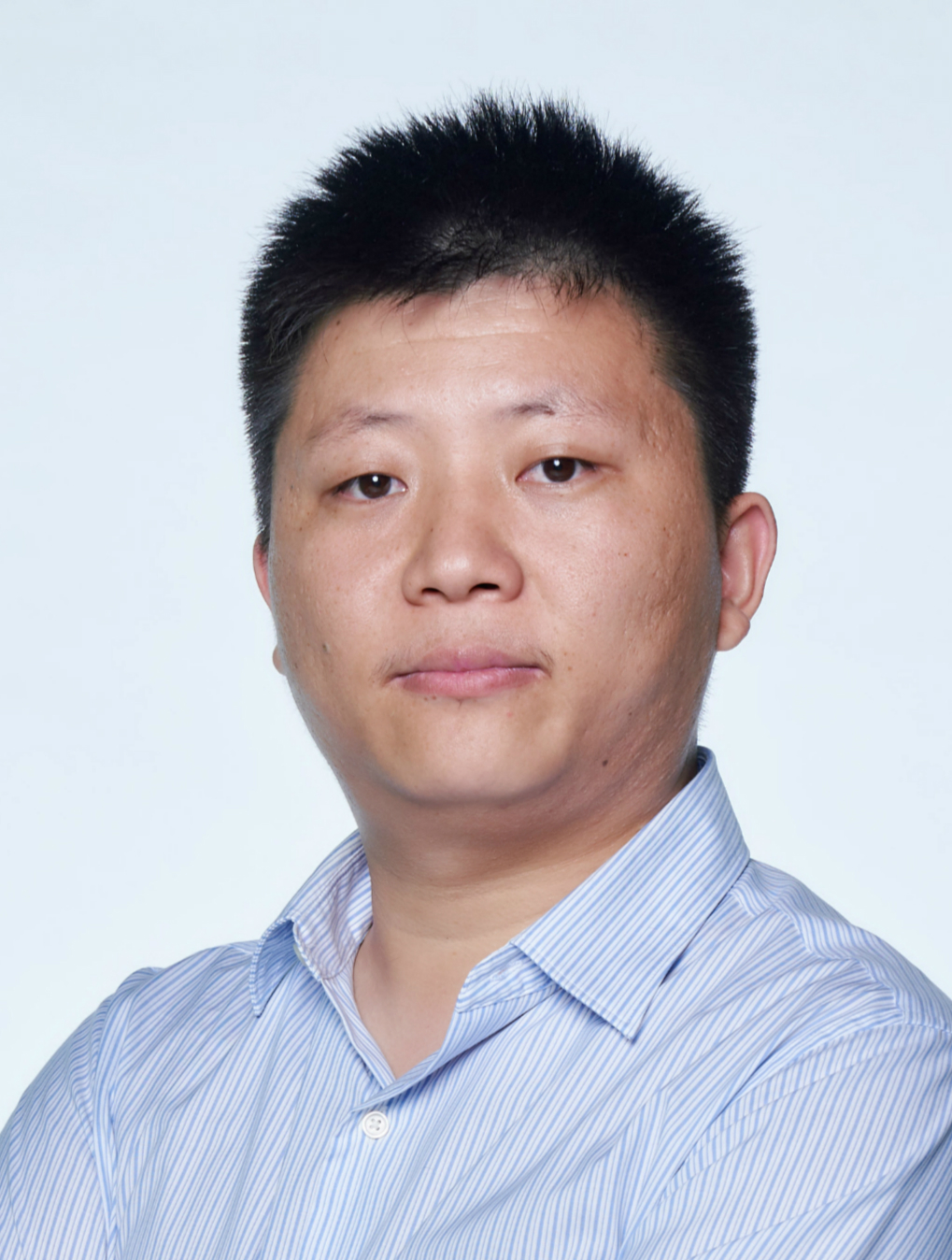}}]
{Xiaofei Xie} is a Wallenberg-NTU Presidential Postdoctoral Fellow  in Nanyang Technological University, Singapore. He received Ph.D, M.E. and B.E. from Tianjin University. His research mainly focus on program analysis, loop analysis, traditional software testing and security analysis of artificial intelligence. He has published some top tier conference/journal papers relevant to software analysis in ISSTA, FSE, TSE, ASE, ICSE, TIFS, TDSC, IJCAI, AAAI, CCS. In particular, he won two ACM SIGSOFT Distinguished Paper Awards.
\end{IEEEbiography}

\begin{IEEEbiography}[{\includegraphics[width=1in,height=1.25in,clip,keepaspectratio]{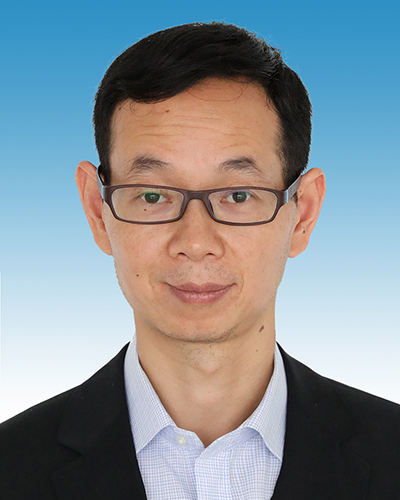}}]
{Shengyong Chen} received the Ph.D. degree from the City University of HongKong, Hong Kong, in 2003. He is a Professor and a Ph.D. Advisor withthe School of Computer Science and Engineering, Tianjin University of Technology, Tianjin, China. He has authored over 100 scientific papers in international journals and conferences. His current research interests include evolutionary computation and intelligent systems.Dr. Chen achieved the National Outstanding Youth Fund of China, in 2013. He is an IET Fellow.
\end{IEEEbiography}

\begin{IEEEbiography}[{\includegraphics[width=1in,height=1.25in,clip,keepaspectratio]{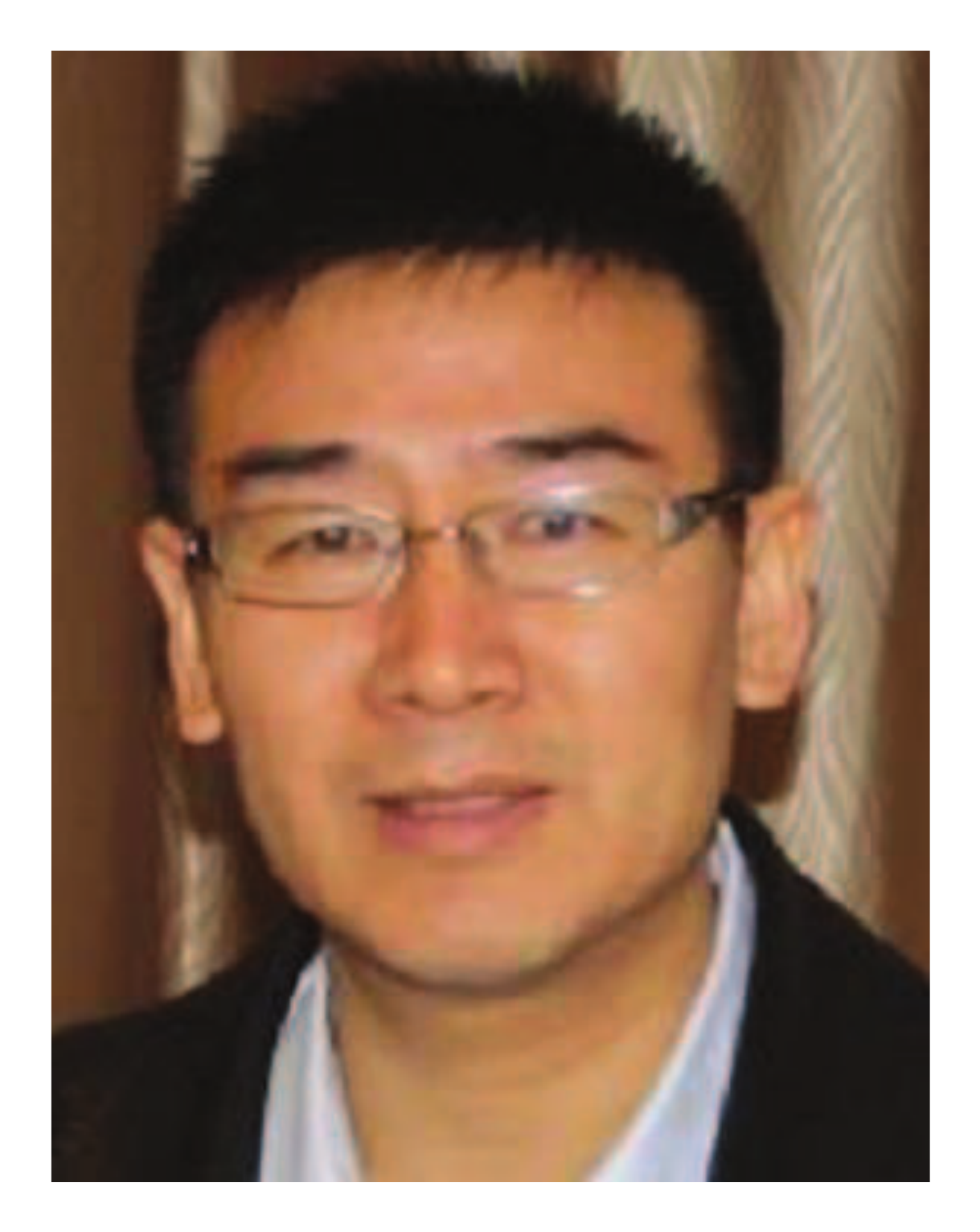}}]
{Rong Chen} He received the M.S. and Ph.D. degree in computer software and theory from the Jilin University, China, in 1997 and 2000. He is currently a professor of the College of Information Science and Technology at the Dalian Maritime University, and has previously held position at Sun Yat-sen University, China. His research interests are in software diagnosis, collective intelligence, activity recognition, Internet and mobile computing. He is a member of the IEEE and a member of the ACM.
\end{IEEEbiography}




\end{document}